\documentclass{article}
\usepackage{arxiv}
\usepackage{mathtools,amssymb, bm}
\usepackage[ruled,vlined]{algorithm2e}
\usepackage[noend]{algpseudocode}
\usepackage{amsfonts}
\usepackage{subcaption}
\usepackage{float}
\usepackage{hyperref}
\usepackage{xcolor}
\usepackage{lineno}
\usepackage{amsthm}
\usepackage{amsmath}
\usepackage{mathrsfs}

\usepackage{bm}
\usepackage{ragged2e}
\justifying
\usepackage{amsmath}
\usepackage{graphicx}
\graphicspath{{{./}}}

\usepackage{amsmath}
\usepackage{amsfonts}
\usepackage{amssymb}

\newcommand{\etal}{{\it et al.}}
\newcommand{\ie}{{ i.e.}}

\usepackage{xcolor}

\newcommand{\Aref}[1]{Alg.\,~\ref{#1}}

\newcommand{\fref}[1]{Fig.\,~\ref{#1}}

\newcommand{\eref}[1]{Eq.\,~(\ref{#1})}

\newcommand{\sref}[1]{Sec.\!~\ref{#1}}

\newcommand{\cref}[1]{Ref.\,~\cite{#1}}

\newcommand{\aposteriori}{{\it{a posteriori}} }

\newcommand{\pdf}{{\pi}}

\newcommand{\bs}{\mathsf{b}}

\newcommand{\gs}{\mathsf{g}}

\newcommand{\ys}{\mathsf{y}}

\newcommand{\Vs}{\mathsf{V}}
\newcommand{\Ws}{\mathsf{W}}

\newcommand{\gb}{\mathbf{g}}

\newcommand{\xb}{\mathbf{x}}

\newcommand{\Hb}{\mathbf{H}}
\newcommand{\Fb}{\mathbf{F}}

\newcommand{\Pb}{\mathbf{P}}

\newcommand{\Ib}{\mathbf{I}}

\newcommand{\Ac}{\mathcal{A}}
\newcommand{\Gc}{\mathcal{G}}
\newcommand{\Ec}{\mathcal{E}}
\newcommand{\Nc}{\mathcal{N}}

\newcommand{\Uc}{\mathcal{U}}
\newcommand{\Sc}{\mathcal{S}}

\newcommand{\Ebb}{\mathbb{E}}

\newcommand{\mub}{\boldsymbol{\mu}}

\newcommand{\epsilonb}{{\boldsymbol{\epsilon}}}

\newcommand{\Sigmab}{\boldsymbol{\Sigma}}

\newcommand{\phib}{{\boldsymbol{\phi}}}

\newcommand{\grad}{{\boldsymbol{\nabla}}}
\newcommand{\partialb}{{\boldsymbol{\partial}}}

\newcommand{\tr}{\operatorname{tr}}

\newcommand{\argmin}{\operatorname{argmin}}

\newcommand{\dt}{\mathrm{d}t}

\newcommand{\NN}{\mathsf{N}\!\mathsf{N}}
\newcommand{\weight}{\mathsf{w}}
\newcommand{\parameters}{{\boldsymbol{\theta}}}
\newcommand{\inputvector}{\mathsf{x}}
\newcommand{\outputvector}{\mathsf{y}}
\newcommand{\hiddenvector}{\mathsf{h}}

\newcommand{\weightmatrix}{\mathsf{W}}
\newcommand{\biasvector}{\mathsf{b}}
\newcommand{\activation}{a}

\newcommand{\prob}{\pi}

\newcommand{\posterior}{\pi}

\newcommand{\giv}{\, | \,}

\newcommand{\inputs}{\mathsf{x}}
\newcommand{\outputs}{\mathsf{y}}
\newcommand{\data}{\mathsf{D}}

\newcommand{\potential}{\Phi}

\newcommand{\kernel}{\kappa}

\newcommand{\KLdiv}{\operatorname{KL}}

\newcommand{\CDF}{\operatorname{CDF}}

\newcommand{\stress}{\mathbf{S}}
\newcommand{\strain}{\mathbf{E}}

\newcommand{\graph}{\mathcal{G}}

\date{}

\title{{\bf Condensed Stein Variational Gradient Descent for Uncertainty Quantification of Neural Networks}}

\author{
{\bf Govinda Anantha Padmanabha} \\
{\it Sibley School of Mechanical and Aerospace Engineering}, \\ 
Cornell University, Ithaca, NY 14850 \\[0.1in]
\And
{\bf Cosmin Safta} \\
{{\it Sandia National Laboratories}},\\
Livermore, CA 94551 \\[0.1in]
\And
{\bf Nikolaos Bouklas} \\
{\it  Sibley School of Mechanical and Aerospace Engineering}, \\ {\it  \& Center for Applied Mathematics,}  \\
Cornell University, Ithaca, NY 14850 \\ Pasteur Labs, Brooklyn, NY 11205 \\[0.1in]
\And
{\bf Reese E. Jones}$^*$ \\
{\it Sandia National Laboratories}, \\
Livermore, CA 94551 \\
$^*$ {Corresponding author: {\tt rjones@sandia.gov}} 
}

\begin{document}
\maketitle

\begin{abstract}

We propose a Stein variational gradient descent method to concurrently sparsify, train, and provide uncertainty quantification of a complexly parameterized model such as a neural network.
It employs a graph reconciliation and condensation process to reduce complexity and increase similarity in the Stein ensemble of parameterizations.
Therefore, the proposed condensed Stein variational gradient (cSVGD) method provides uncertainty quantification on parameters, not just outputs.
Furthermore, the parameter reduction speeds up the convergence of the Stein gradient descent as it reduces the combinatorial complexity by aligning and differentiating the sensitivity to parameters.
These properties are demonstrated with an illustrative example and an application to a representation problem in solid mechanics.
\end{abstract}

{\it 
Bayesian neural network, graph neural network, Stein variational inference, sparsification, graph
condensation, uncertainty quantification.}

\section{Introduction}

In the context of uncertainty quantification (UQ) the \emph{curse of dimensionality}, whereby quantification efficiency degrades drastistically with parameter dimension, is particular extreme with highly parameterized models such as neural networks (NNs).
Fortunately, in many cases, these models are overparameterized in the sense that the number of parameters can be reduced with negligible effects on accuracy and sometimes improvements in generalization~\cite{fuhg2024extreme}.
Furthermore, NNs often have parameterizations that have fungible parameters such that permutations of the values and connections lead to equivalent output responses.
This suggests methods that simultaneously sparsify and characterize the uncertainty of a model, while handling and taking advantage of the symmetries inherent in the model, are potentially advantageous approaches.

Although Markov chain Monte Carlo (MCMC) methods~\cite{ghanem2017handbook} have been the reference standard to generate samples for UQ methods, they can be temperamental and do not scale well for high dimensional models.
More recently, there has been widespread use of variational inference methods (VI), which cast the parameter posterior sampling problem as an optimization of a surrogate posterior guided by a suitable objective, such as the Kullback-Liebler (KL) divergence between the predictive posterior and true posterior induced by the data.
In particular, there is now a family of model ensemble methods based on Stein's identity~\cite{stein1987large}, such as Stein variational gradient descent (SVGD)~\cite{liu2016stein}, projected SVGD~\cite{chen2019projected}, and Stein variational Newton's method~\cite{detommaso2018stein}.
These methods have advantages over MCMC methods by virtue of propagating in parallel a coordinated ensemble of particles that represent the empirical posterior.

In our recent previous work~\cite{padmanabha2024improving}, we proposed a sequential sparsification-uncertainty quantification method based on $L_{0}$ regularization and Stein gradient flow.
We showed that sparsification by $L_0$ methods~\cite{louizos2017learning} was able to capture the nonlinear nature of the reduced parameter manifold, which subsequently was explored by a Stein ensemble of particles representing likely model parameterizations.
The sequential application of sparsification and then SVGD displayed advantages over established methods~\cite{chen2019projected} on a number of physical representation problems.

The {\it concurrent} sparsification-uncertainty quantification method presented in this work, termed cSVGD, has a number of advantages over our sequential method.
Foremost, the new method enables parameter alignment and leads to more interpretable parameter UQ, versus solely providing output uncertainty estimates.
This also enables the Stein method to avoid spurious repulsion or lack of repulsion due to parameter permutations.
Furthermore, there is an improvement in computational speed due to the reduction in the complexity of the models in the Stein ensemble.
Parameter UQ and computational speed-ups are enabled by reconciling the parameters by importance across the Stein ensemble and condensing the common representation of the model as the gradient flow and sparsification progress.
The method has a firm theoretical foundation based on the introduction of sparsifying priors into the Stein variational gradient flow formalism~\cite{liu2016stein}.

In the next section, \sref{sec:related}, we give an overview of related work that we build upon.
Then, in  \sref{sec:methods}, we develop the concurrent sparsifying Stein gradient descent (cSVGD) algorithm for feedforward neural networks (FFNNs). In \sref{sec:demonstration}, we illustrate the general behavior of the algorithm, in particular, the interplay of the hyperparameters controlling the prior and the Stein kernel. In \sref{sec:results}, we demonstrate the performance of the method on a problem of scientific interest: determining a hyperelastic potential from stress-strain data. Finally, in \sref{sec:conclusion}, we conclude with a summary of findings and avenues for future work.

\section{Related work} \label{sec:related}

To provide an efficient alternative to MCMC sampling
Liu and Wang~\cite{liu2016stein,liu2017stein} introduced  Stein variational gradient descent (SVGD) which propagates a coordinated ensemble of parameter realizations that approximate the posterior distribution.
Newton~\cite{detommaso2018stein,leviyev2022stochastic} and projected~\cite{chen2019projected} variants soon followed.
Our previous work~\cite{padmanabha2024improving} combined sparsification and SVGD in sequence to fully and efficiently explore the submanifold of influential parameters.
That work relied on $L_p$, $p \in [0,1]$, regularizations that induced parameter sparsity, such as the work by Louizos \etal~\cite{louizos2017learning} which developed a smoothed $L_0$ regularization.
Others, such as McCulloch \etal~\cite{mcculloch2024sparse}, have used this type of regularization in scientific machine learning.
Another notable example of sparse parameter identification is the Sparse Identification of Nonlinear Dynamical systems (SINDy) framwork~\cite{brunton2016discovering}.
In a probabilistic, Bayesian context these regularizations have analogs in sparsifying priors, such as the Cauchy prior introduced in Bayesian compressive sensing and Least Absolute Shrinkage and Selection Operator (LASSO)~\cite{tibshirani1996regression,ji2008bayesian,babacan2009bayesian,baron2009bayesian}.

Our augmentation of sparsifying priors in the Stein flow with model graph condensation to reduce complexity has some similarity to the well-established field of active set methods~\cite{lewis2002active,ferreau2008online,wen2012convergence,misra2022learning}.
These methods rely on transformations and/or heuristics to reduce the permutational complexity of identifying the active constraints in a large set of inequality constraints~\cite{misra2022learning} or the optimum in a non-smooth problem \cite{wen2012convergence}.

\section{Concurrent sparsification and Stein variational gradient descent} \label{sec:methods}
First, we develop Stein variational gradient descent (SVGD) with a sparsifying prior as the basis for the proposed algorithm. Then, we show how we achieve parameter alignment by treating the Stein ensemble as an ensemble of graphs with a common graph. Finally, we summarize the overall algorithm.
\subsection{Bayesian parameter uncertainty quantification}
Given a model $\hat{\outputs}(\inputs; \parameters)$ with inputs $\inputs$, parameters $\parameters = (\theta_1, \theta_2,\ldots,\theta_n)$, and data $\data = \{ \inputs_i, \outputs_i \}_{i=1}^{N_D}$, where $N_D$ is the number of data samples; Bayes rule expressed as:
\begin{equation} \label{eq:bayes}
\prob(\parameters \giv \data) = \frac{\prob(\data \giv \parameters) \prob(\parameters)} { \prob(\data) }
\end{equation}
provides the posterior distribution $\prob(\parameters \giv \data)$ on the parameters $\parameters$ given data $\data$.
The formulation is also dependent on choices for the likelihood $\prob(\data \giv \parameters)$ and the prior distribution of parameters $\prob(\parameters)$.
These distributions determine the posterior and the evidence $\prob(\data)$.

\begin{figure}
\centering
\includegraphics[width=0.75\linewidth]{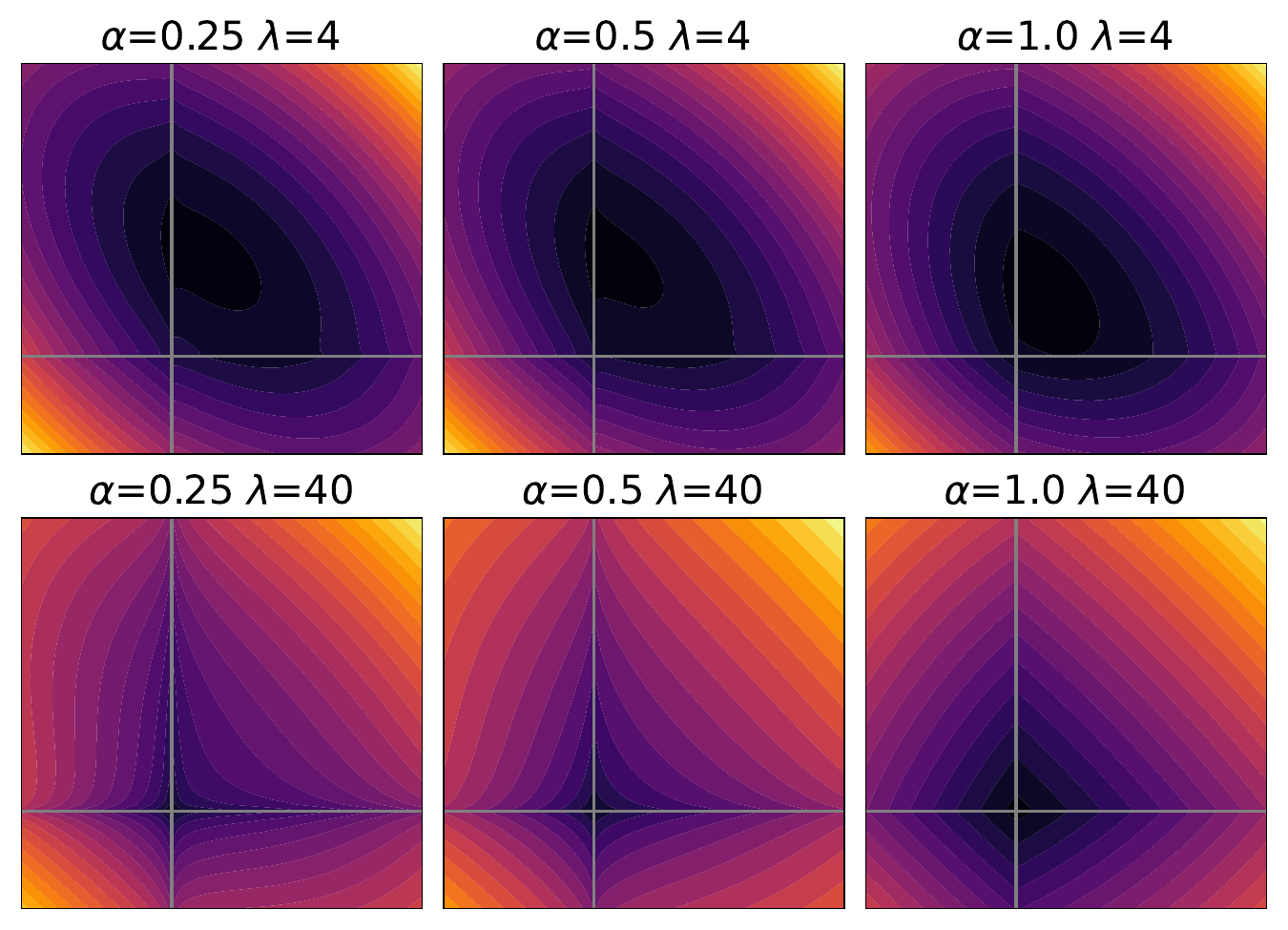}
\caption{Log posteriors formed from a Gaussian likelihood and priors from the $\alpha$ exponential family scaled by $\lambda$.
As $\lambda$ is increased (top to bottom), the posterior resembles the prior mode more than the likelihood, and as $\alpha$ is reduced (right to left), the contours of the prior transform from diamond-shaped to more cruciform-shaped.
Note the values of $\lambda$ are relative to the particular likelihood chosen for this illustration.
}
\label{fig:prior_family}
\end{figure}

For example, with the assumption of a Gaussian likelihood
\begin{equation}
\prob(\data \giv \parameters) \propto \exp\left( -\frac{1}{2 \sigma^2} \left| \ys - \hat{\ys}(\inputs; \parameters) \right|^2 \right)
\end{equation}
and a sparsifying prior
\begin{equation} \label{eq:prior_family}
\prob(\parameters) \propto   \exp(-\lambda \left|\parameters \right|^\alpha)
\quad \alpha \in (0,1]
\end{equation}
the negative log posterior becomes
\begin{equation} \label{eq:log_post}
-\log \prob(\parameters \giv \data)
= \frac{1}{2\sigma^2}\left| \ys - \hat{\ys}(\inputs; \parameters) \right|^2
+ \lambda \left| \parameters \right|^\alpha + ... ,
\end{equation}
where $\sigma^2$ is the data noise variance and $\lambda$ acts as a penalty parameter on the regularizating objective $\left| \parameters \right|^\alpha$.
Note the remaining additive terms in \eref{eq:log_post} do not depend on the parameters $\parameters$ and, hence, do not affect the gradient of $\log \prob(\parameters \giv \data)$, otherwise known as a the \emph{score}.
Here, and in the following, we use $\left| \parameters \right|^\alpha$ to denote
\begin{equation}
\left| \parameters \right|^\alpha
= \sum_i \left| \theta_i \right|^\alpha
\end{equation}
i.e. $\left| \parameters \right|^\alpha$  is related to the $L_\alpha$ norm by $\left| \parameters \right|^\alpha = (\| \parameters \|_\alpha)^{\alpha}$.
\fref{fig:prior_family} illustrates this family of posteriors; notice that as $\alpha\to 0$ the prior approximates the $L_0$ prior.

Specifically, we employ priors in the exponential family
\begin{equation}
\prob(\parameters) = \lambda {c_1(\alpha)} \exp \left(-\lambda^\alpha {c_2(\alpha)}\left|\parameters\right|^\alpha\right),
\label{eq:sparprior}
\end{equation}
where
\begin{equation}
c_1(\alpha)=\frac{\alpha \Gamma(3 / \alpha)^{1 / 2}}{2 \Gamma(1 / \alpha)^{3 / 2}}  \quad \text{and} \quad
c_2(\alpha)=\left[\frac{\Gamma(3 / \alpha)}{\Gamma(1 / \alpha)}\right]^{\alpha / 2} \ .
\end{equation}
and $\Gamma$ is the Gamma function.

\subsection{Variational inference}
As an alternative to sampling the posterior distribution with standard but computationally expensive Markov chain Monte Carlo (MCMC) methods \cite{geyer2011introduction}, variational inference (VI) reformulates the task of approximating the posterior as an optimization problem on the parameters of a surrogate posterior.
In VI, the optimization problem for the surrogate $q(\parameters)$ is typically based on the Kullback-Liebler (KL) divergence:
\begin{eqnarray} \label{eq:surrogate_dist}
q^{*}(\parameters)
&=&\underset{\varphi}{\arg \min }\, {\KLdiv}(q(\parameters; \varphi) \| \prob(\parameters \giv \data))
\\
&=&\underset{\varphi}{\arg \min}\, \mathbb{E}_{q}[\log q(\parameters; \varphi)-\log p (\parameters | \mathcal{D})] \ , \nonumber
\end{eqnarray}
measuring the difference between the surrogate $q$ and true posterior $\prob(\parameters \giv \data)$ associated with the data $\data$.
Given that $p(\parameters \giv \data)$ appears in a logarithm in \eref{eq:surrogate_dist}, the evidence does not need to be evaluated (since the it is a constant) in the search for the optimum $q^{*}$.

The SVGD algorithm \cite{liu2016stein} is a VI method based on the Stein identity \cite{stein1987large}
\begin{align}
\Ebb_{\parameters\sim\pi}[ \Sc_\prob [ \phib(\parameters) ] ] = \mathbf{0},
\end{align}
where
\begin{align}
\Sc_\prob [\phib(\parameters)] = \grad_\parameters [ \log \prob(\parameters)] \otimes \phib(\parameters) + \grad_\parameters \phib(\parameters)
= \frac{\grad_\parameters [  \prob(\parameters)\phib(\parameters) ] }{ \prob(\parameters)}
\end{align}
for a smooth function $\phib$ and a generic probability density $\pdf(\parameters)$.
The Stein operator $\Sc_{\prob}$ depends on the density $\prob$ only through the \emph{score} function $\grad_\parameters [\log \prob(\parameters)]$ which is independent of the normalization of $\pdf$ and thus avoids the evaluation of the evidence.

If we let $q$ be an approximate density under a perturbation mapping of the random variable $\parameters \mapsto \parameters+\epsilon \phib$, then the expectation of $\tr (\Ac_\prob [ \phib (\parameters)])$ is related to the gradient of the Kullback--Leibler (KL) divergence between $q$ and $\prob$~\cite[Thm 3.1]{liu2016stein}:
\begin{align}
\left.
\grad_\varepsilon D_{\text{KL}}( q_{\parameters+\epsilon \phib} ||  \prob)
\right|_{\varepsilon=0}
= - \Ebb_{q} [ \tr(\Ac_\prob [ \phib(\parameters) ] ].
\end{align}
This provides the basis for a gradient-based method to update the approximate posterior density.

The SVGD algorithm utilizes an ensemble of $N_r$ \emph{particles} $\{\parameters_{(i)}\}_{i=1}^{N_r}$ in the parameter space.
After randomly initializing the parameter ensemble, SVGD updates the particles via gradient-descent steps
\begin{align}
\parameters^{(k+1)}_{a} = \parameters^{(k)}_{a} + \epsilon \gs({\parameters^{(k)}_{(i)}}) \ ,
\end{align}
where $\epsilon$ is a step size parameter, $k$ is the iteration counter, and
\begin{eqnarray}
\gs(\parameters)
&=& \frac{1}{N_r}\sum_{b=1}^{N_r}
\frac{\grad_{\parameters^{(k)}_b} \left(
\kernel(\parameters^{(k)}_b, \parameters) \posterior(\parameters^{(k)}_b \giv \data) \right)}
{\posterior(\parameters^{(k)}_b \giv \data)}
\label{eq:pseudopotential} \\
&=& \frac{1}{N_r}\sum_{b=1}^{N_r} \left[
\kernel(\parameters^{(k)}_b, \parameters)
\frac{\grad_{\parameters^{(k)}_b}
\prob(\data \giv \parameters^{(k)}_b)}
{ \prob(\data \giv \parameters^{(k)}_b) }
+
\frac{\grad_{\parameters^{(k)}_b} \left(
\kernel(\parameters^{(k)}_b, \parameters) \prob(\parameters^{(k)}_b) \right)}
{\prob(\parameters^{(k)}_b)}  \right]
\label{eq:grad_split} \\
&=&
\frac{1}{N_r}\sum_{b=1}^{N_r} \left[
\kernel(\parameters^{(k)}_b, \parameters) \left(
\grad_{\parameters^{(k)}_b}
[\log \prob(\data \giv \parameters^{(k)}_b ) ]
+
\grad_{\parameters^{(k)}_b}
[\log \prob(\parameters^{(k)}_b ) ] \right) \right.
\nonumber \\ &&
+
\left.\grad_{\parameters^{(k)}_b}
\kernel(\parameters_{b}^{(k)}, \parameters) \right]
\label{eq:stein_grad}
\end{eqnarray}
is the parameter gradient which depends on the choice of the kernel $\kernel(\cdot,\cdot)$.
Note the sum is over all particles and $\grad_{\parameters} \kernel(\parameters,\parameters) \equiv \mathbf{0}$ by definition.
We will call the kernel smoothed posterior $\kernel(\parameters^{(k)}_b, \parameters) \posterior(\parameters^{(k)}_b \giv \data)$ apparent in \eref{eq:pseudopotential}  the \emph{pseudopotential}.

In the form of the gradient in \eref{eq:grad_split} we see that the first term depends solely on the likelihood and kernel, while the second term depends only on the prior and the kernel.
While, in the form of the gradient in \eref{eq:stein_grad}, the first term emphasizes data fit through the likelihood, the second term encourages sparsification through the prior, and the third term penalizes similar parameterizations through the repulsive kernel.

More specifically, the terms in \eref{eq:stein_grad} can be interpreted as: (a) a kernel-smoothed gradient that drives the particles toward the best-fit region according to the likelihood, (b) a sparsifying force whose magnitude depends on $\lambda$ that is present in the definition of the prior density in \eref{eq:sparprior}, and (c) a repulsion term that keeps the particles distinct.
For the second term, the scaling is explicit in $\lambda$, but for the third term, it is implicit in the bandwidth of the kernel.
A traditional Stein flow with an uninformative prior involves only (a) and (c), i.e., $\lambda = 0$, while a gradient flow to find a sparse maximum \aposteriori (MAP) point involves only (a) and (b) with the bandwidth $\gamma \to \infty$ such that the kernel is flat and the gradient in (c) is zero.
Depending on the number of fungible parameters and the number of negligible parameters, the sparsification and repulsion terms can conflict with their objectives, which we will explore in the next section, \sref{sec:demonstration}.

We use an exponential family of kernels similar to the prior family described in \eref{eq:prior_family}, where a sparsifying prior of the form $\prob(\parameters) \propto   \exp(-\lambda \left|\parameters \right|^\alpha), \alpha \in (0,1]$ was utilized:
\begin{equation}
\kernel(\parameters',\parameters) = \exp\left(-\frac{1}{\gamma \beta} | \parameters' - \parameters |^\beta\right) ,
\end{equation}
with the idea the that lower $\beta$ exponential kernels will repel more strongly along coordinate directions and potentially add sparsification.
These kernels have the gradient
\begin{equation}
\grad_{\parameters'} \kernel(\parameters', \parameters)
=
\frac{1}{\gamma} | \parameters - \parameters'|^{\beta-1} \operatorname{sign}(\parameters - \parameters') \kernel(\parameters', \parameters)
\end{equation}
For this family,  the second term in \eref{eq:grad_split} simplifies to
\begin{equation} \label{eq:prior_grad}
\frac{\grad_{\parameters^{(k)}_b} \left(
\kernel(\parameters^{(k)}_b, \parameters) \prob(\parameters^{(k)}_b) \right)}
{\prob(\parameters^{(k)}_b)}
=
\exp(-\gamma | \parameters^{(k)}_b - \parameters |^{\beta})
\grad_{\parameters^{(k)}_b} \left(
-\gamma | \parameters^{(k)}_b- \parameters |^{\beta} + \lambda | \parameters^{(k)}_b |^{\alpha} \right)
\end{equation}
using the identity that the score of a product is the sum of the scores
\begin{equation}
\frac{\grad_\xb (a(\xb) b(\xb)))}{ a(\xb) b(\xb) } =  \frac{ \grad_\xb a(\xb) }{ a(\xb) } + \frac{\grad_\xb b(\xb) }{  b(\xb) } \ .
\end{equation}
So, in this manner, the (kernel smoothed) gradient of the log kernel and log prior directly affect the Stein flow, which is guided by the data through the likelihood.

\subsection{Graph condensation}

We focus on complex models, such as neural networks, with many parameters, some of which are redundant or insignificant.

The FFNNs model have a structure based on layers $\ell$ composed of nodes.
The output is the composition of trainable affine transforms $\Ws_\ell \cdot + \bs_\ell$ and preselected element-wise nonlinear transforms $a_\ell(\cdot)$:
\begin{equation}
\outputs = \NN(\inputs; \parameters) = \activation_{N_\ell}( \weightmatrix_{N_\ell}(\ldots \underbrace{\activation_0(\weightmatrix_0 \inputs + \biasvector_0)}_{\outputs_{1}}\ldots \biasvector_{N_\ell})
\end{equation}
where $\outputs_1$ is the output of the first layer ($\ell$=1) and the parameters of the model are $\parameters = \{\weightmatrix_\ell, \biasvector_\ell \}_{\ell=1}^{N_\ell}$  consisting of weight matrices $\weightmatrix_\ell$ and bias vectors $\biasvector_\ell$.
Due to the fungibility of parameters in a layer, multiple permutations of the same parameters produce identical output.
In fact, nothing in the (regularized or unregularized) optimization process breaks this degeneracy.
In an ensemble of parameter realizations, forward propagation is unhindered but uncertainties of parameters are confounded by this permutational symmetry, in the sense that parameters playing similar roles in are not in identical nodes of the NN.
Furthermore, this condensation ameliorates the potential for spurious particle repulsion in the Stein ensemble.

Hence, we reconcile the parameterizations by maximizing their similarity.
First, we represent the NNs as directed graphs
$\Gc = \{\Nc, \Ec \}$ with nodes $\Nc$ associated with biases $\bs_\ell = [b_I]_\ell$ and edges $\Ec$ with weights $\Ws = [W_{IJ}]_\ell$.
In a graph representing an FFNN all nodes in a layer $\ell$ are fungible so that permutations $\Pb_{\ell}$ for all the hidden layers $\ell \in [2,N_\ell-1]$ (\ie\ excluding input $\ell=1$ and output layers $\ell=N_\ell$)
\begin{equation}
\weightmatrix'_\ell = \Pb_{\ell+1}^T  \weightmatrix_\ell \Pb_{\ell}
\quad \text{and} \quad
\biasvector'_\ell =  \Pb_{\ell+1}^T  \biasvector_\ell
\end{equation}
produce models with identical output.
A particular version of the similarity objective
is to find permutations that attain a minimum distance across the ensemble:
\begin{equation}
\argmin_{\{\Pb_a\}}
\| d(\NN_a, \NN_b) \| \quad a \in [1,N_r] \ \text{and} \ b \in [a+1,N_r]
\end{equation}
where the permutations $\Pb_a = \{\Pb_{\ell,a}\}$ for realization $a=1$ are fixed at the identity and the pairwise distance is
\begin{equation} \label{eq:distances}
d(\NN_a, \NN_b) =
\sqrt{
\sum_\ell \| \weightmatrix_{\ell,a} - \weightmatrix_{\ell,b} \|_2^2
} ,
\end{equation}
In \eref{eq:distances} we have ignored biases, which will be appropriate for the model we use in the demonstration but, in general, may not be sufficient to distinguish parameterizations.

In this work, we employ a heuristic algorithm based on sorting and pruning to obtain a solution to the parameter reconciliation problem.
First, we identify and remove edges associated with insignificant (near zero) weights; then we remove disconnected nodes \,  i.e., \ those with only zero incoming or outgoing weights.
Second, we sort the nodes in each layer based on criterion \emph{importance}.
\begin{equation} \label{eq:importance}
s_{\ell,j} = \sum_i [ \weightmatrix_\ell ]_{ij}
\end{equation}
which is the influence of node $j$ in layer $\ell$ on all nodes in the subsequent layer and $[ \Ws_\ell ]_{ij}$ are components of the $\weight_\ell$ weight matrix.
Third we put all graphs in the ensemble on a common graph that is defined by the maximum nodes for each layer across the ensemble.
Graphs that do not use all nodes in this template are padded to fit.
\Aref{alg:graph_condense} summarizes the algorithm and provides more details.

\begin{algorithm}[!ht] \SetAlgoLined
\textbf{Input}: A set of directed graphs $\{ \graph_a \}$ representing FFNN realizations in a ensemble \\
\While{ ensemble $\{ \graph_a \}$ {\rm is not changing topologically}}{
{\bf (1) Prune and sort each graph} \\
\For { graph $\graph \in \{ \graph_a \}$} {
Prune insignificant edges, $[\Ws_\ell]_{ij} < \varepsilon$ \\
Prune nodes that do not have at least one incoming edge and outgoing edge with non-zero weights\\
Sort nodes in each layer by \emph{importance} \eref{eq:importance} based on their outgoing edge weights \\
Reorder nodes in each layer and remap edge connections\\
}
{\bf (2) Determine the common graph}\\
Define common graph template $\bar{\graph}$ as a fully connected layer-to-layer graph with $n_\ell$ nodes in layer $\ell \in [0,N_\ell]$ \\
\For {layer $\ell \in [2,N_\ell-1]$}{
Determine maximum number of active nodes $n_\ell$ of layer $\ell$ across ensemble $\{ \graph_a \}$
}
{\bf (3) Reconcile ensemble}\\
\For { graph $\graph \in \{ \graph_a \}$ } {
Insert $\graph$ into template $\bar{\graph}$ with node padding as needed\\
}
} 
\textbf{Result:} A set of directed graphs $\{ \graph_a \}$ sharing a common graph structure with quasi-optimal edge similarity
\caption{Graph condense}
\label{alg:graph_condense}
\end{algorithm}

\subsection{Algorithm: SVGD with concurrent sparsification.}
The overall algorithm is summarized in \Aref{alg:svgd}.
Beyond the choice of prior order $\alpha$ and kernel order $\beta$, the hyperparameters are:
(a) the initial penalty $\lambda_0$,
(b) the kernel width $\gamma$, and
(c) the $N_r$ number of particles.
The initial penalty should be chosen low enough to only slightly perturb the posterior so as to promote initial likelihood driven convergence.
The kernel width can be chosen empirically, for example, with Silverman's rule~\cite{Silverman:1986}, or adaptively.
The dimensionality of the parameter space and the computational budget guide the choice of the number of parameters.
Paramount is to scale $\lambda$ to achieve a balance between the competing objectives of sparsification and accuracy.

At the convergence of \Aref{alg:svgd}, the particles collectively approximate the posterior distribution empirically:
\begin{equation}
\prob(\parameters \giv \data) \approx q(\parameters)
= \sum_{a=1}^{N_r} \delta(\parameters - \parameters_a)
\end{equation}
where $N_r$ is the number of realizations/particles.
A kernel density estimate (KDE) of the particles enables resampling of the posterior.

\begin{algorithm}[!ht]
\textbf{Input}:
\text{A set of initial particles} $\{\parameters_a^{(0))}\}_{a=1}^{N_r}$,
\text{score function} $\nabla \log \prob(\parameters\mid \data)$,
\text{kernel} $\kernel(\parameters,\parameters^{'})$,
\text{step-size} $\{\epsilon\}$,
\text{initial penalty} $\lambda^{(0)}$ \\
\SetAlgoLined
\For{stage $s$}{
\For{iteration $k$}{
Compute gradient
$\gb_a^{(k)}$
from \eref{eq:stein_grad}\\
Update parameters $\parameters_a^{(k+1)} \ \leftarrow \ \parameters_a^{(k)}+\epsilon \gb(\parameters_a^{(k)})$ \\
\textbf{If} converged \textbf{then} break \\
}
{\bf{compact models}}: clip inactive edges, put all particles on common graph using \Aref{alg:graph_condense} \\
{\bf{adapt sparsifying penalty} $\lambda$}:
if accuracy (MSE) has not decreased from previous stage increase $\lambda$, else terminate
\\
}
at termination, revert to minimal $\lambda$ to remove the effect of the prior and increase accuracy with the fixed model form.\\
\textbf{Result:}
A set of particles $\{ \parameters_a \}_{a=1}^{N_r}$ that approximates the target posterior
\caption{Stein variational gradient descent with concurrent sparsification.}

\label{alg:svgd}
\end{algorithm}

\section{Illustrative example} \label{sec:demonstration}

We illustrate the behavior of the Stein ensemble under the combined influence of a Gaussian likelihood, a sparsifying $\alpha$ exponential prior and a repulsive $\beta$ exponential kernel.
Specifically, the chosen likelihood is a multvariate normal (MVN) with mean $\mub$ and precision $\Sigmab^{-1}$:
\begin{equation}
\mub = (1,2,3)  \quad \text{and} \quad
\Sigmab^{-1} =
\begin{bmatrix}
2 & 1 & 0 \\
1 & 2 & 0 \\
0 & 0 & 0.0025 \\
\end{bmatrix} \ .
\end{equation}
The  $\alpha$ prior is associated with the multiplier $\lambda$, and the $\beta$ kernel with bandwidth $\gamma$.
An ensemble of $N_\text{particles} = 128$ particles is more than sufficient to recover the posterior mean and covariance structure in the 3 dimensional parameter space.
Given the low precision of $\theta_3$, we expect the algorithm to sparsify $\theta_3$.
Note by sparsifying $\theta_3$ we can not expect a good estimate of the covariance for that parameter, which illustrates the competing sparsfication and repulsion objectives.

We measure error in the predicted posterior using distance between the empirical mean and covariance of the particle ensemble and true moments of the MVN.
Specifically, we use the Bhattacharyya distance
\begin{equation}
d(\Nc', \Nc) =
\frac{1}{8} (\mub' - \mub) \bar{\Sigmab}^{-1} (\mub' - \mub)
+ \frac{1}{2} \log\left( \frac{\det \bar{\Sigmab}}{\sqrt{\det \Sigmab' \det \Sigmab}} \right)
\end{equation}
where
$\bar{\Sigmab} = \frac{1}{2} (\Sigmab' + \Sigmab)$,
which is sensitive to discrepancies in the mean and some aspects of the covariance.
We measure sparsity with a $L_1$ norm of $\theta_3$ across the ensemble.

\begin{figure}
\centering
\includegraphics[width=0.32\linewidth]{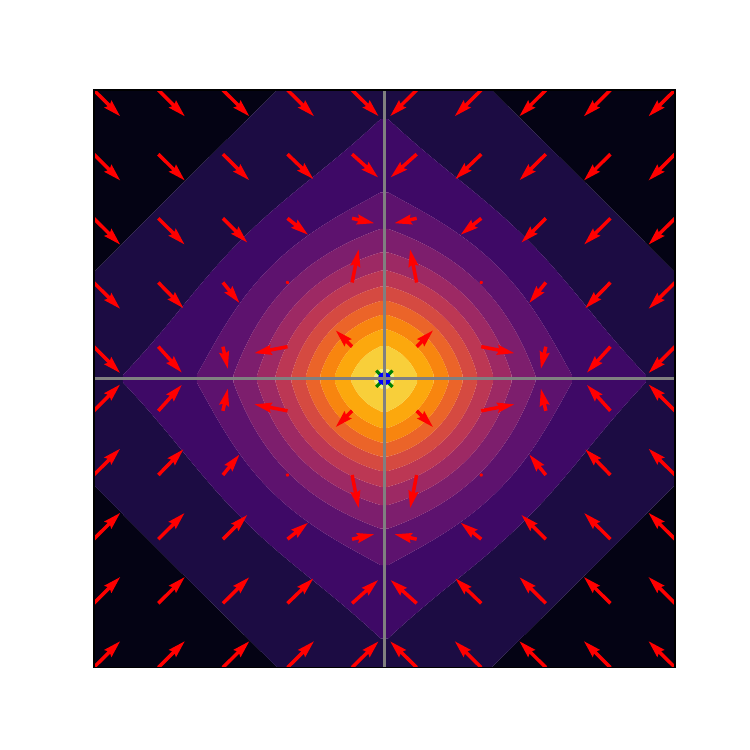}
\includegraphics[width=0.32\linewidth]{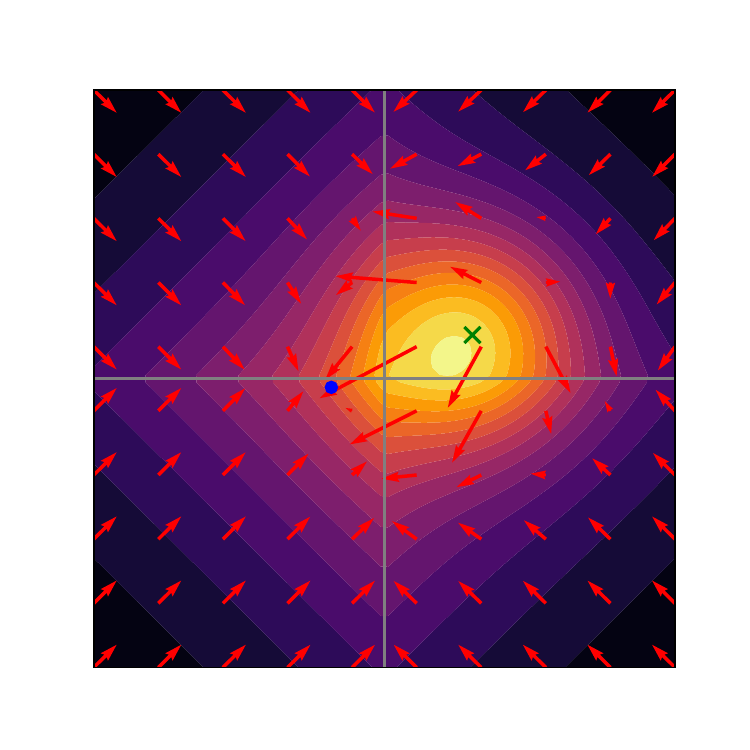}
\includegraphics[width=0.32\linewidth]{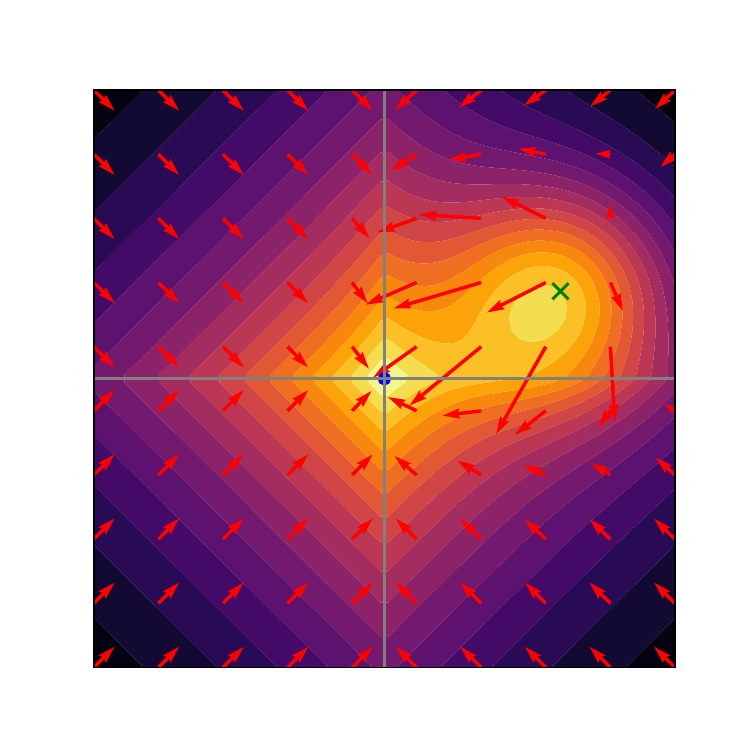}
\caption{Illustration of the Stein gradient (red arrows) and pseudopotential (contours) for $\alpha=1$ prior and $\beta=2$ kernel, one particle (blue dot) in the field of another (green X) which is fixed:
(left) fixed particle is at the origin,
(middle) fixed particle is near the origin,
(right) fixed particle is away from the origin.
The multiplier $\lambda = 1$ and the bandwidth $\gamma = 1/10$.
}
\label{fig:mvn_gradient}
\end{figure}

\fref{fig:mvn_gradient} shows, in the absence of data, how a particle, under the influence of the kernel effects of another particle, is perturbed only if the other particle is also close to the origin.
In this illustration,  the gradient is only from the prior, as in \eref{eq:prior_grad}, and the contour plot depicts the pseudopotential corresponding to $\alpha=1$, $\beta=2$, and $\gamma=1$.
For practical demonstration in the next section, \sref{sec:results}, we modify \Aref{alg:svgd} to mask kernel effects when the particles are close to parameter axes.

\begin{figure}
\centering
\begin{subfigure}[c]{0.95\linewidth}
\includegraphics[width=0.95\linewidth]{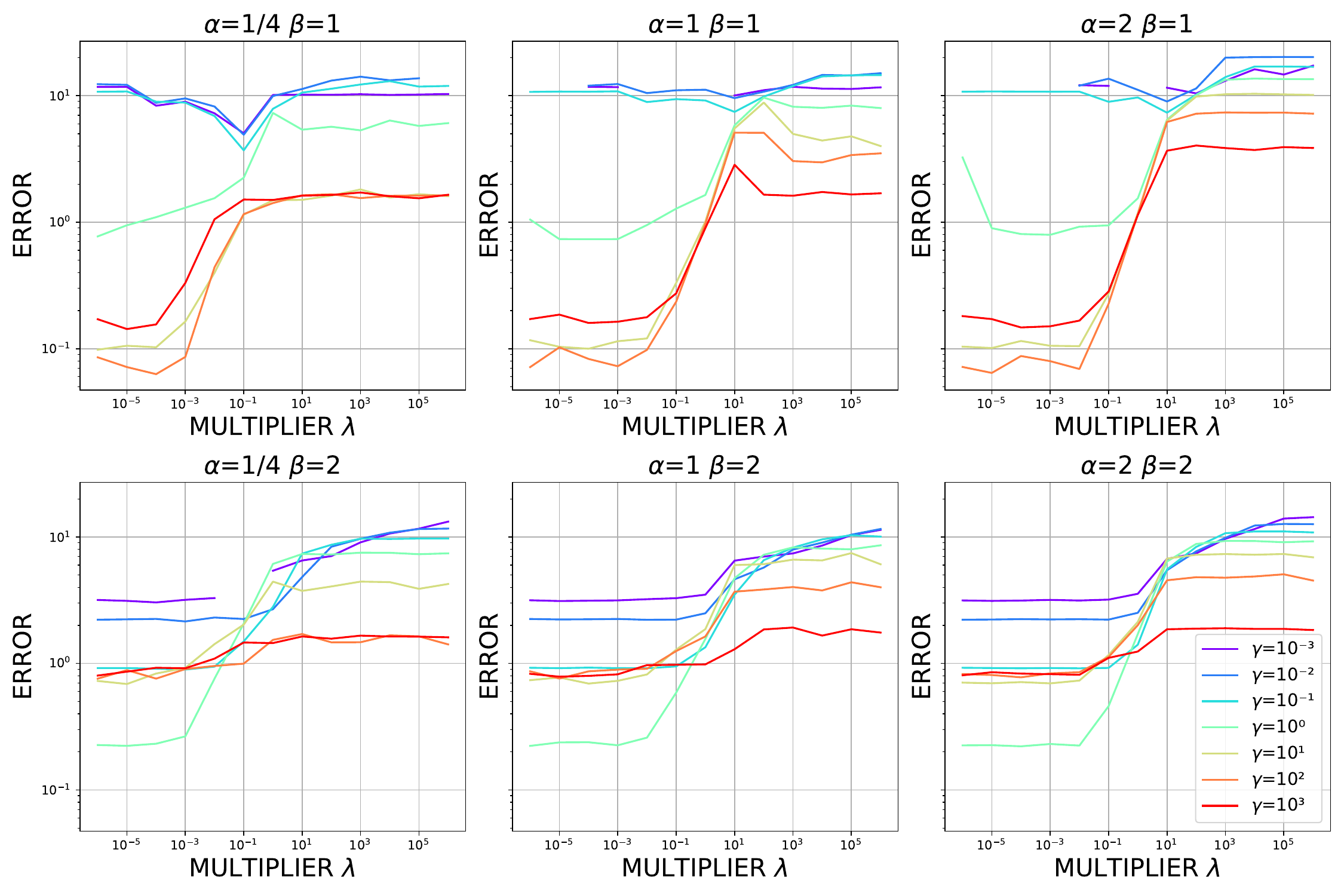}
\caption{Posterior error}
\end{subfigure}
\begin{subfigure}[c]{0.95\linewidth}
\includegraphics[width=0.95\linewidth]{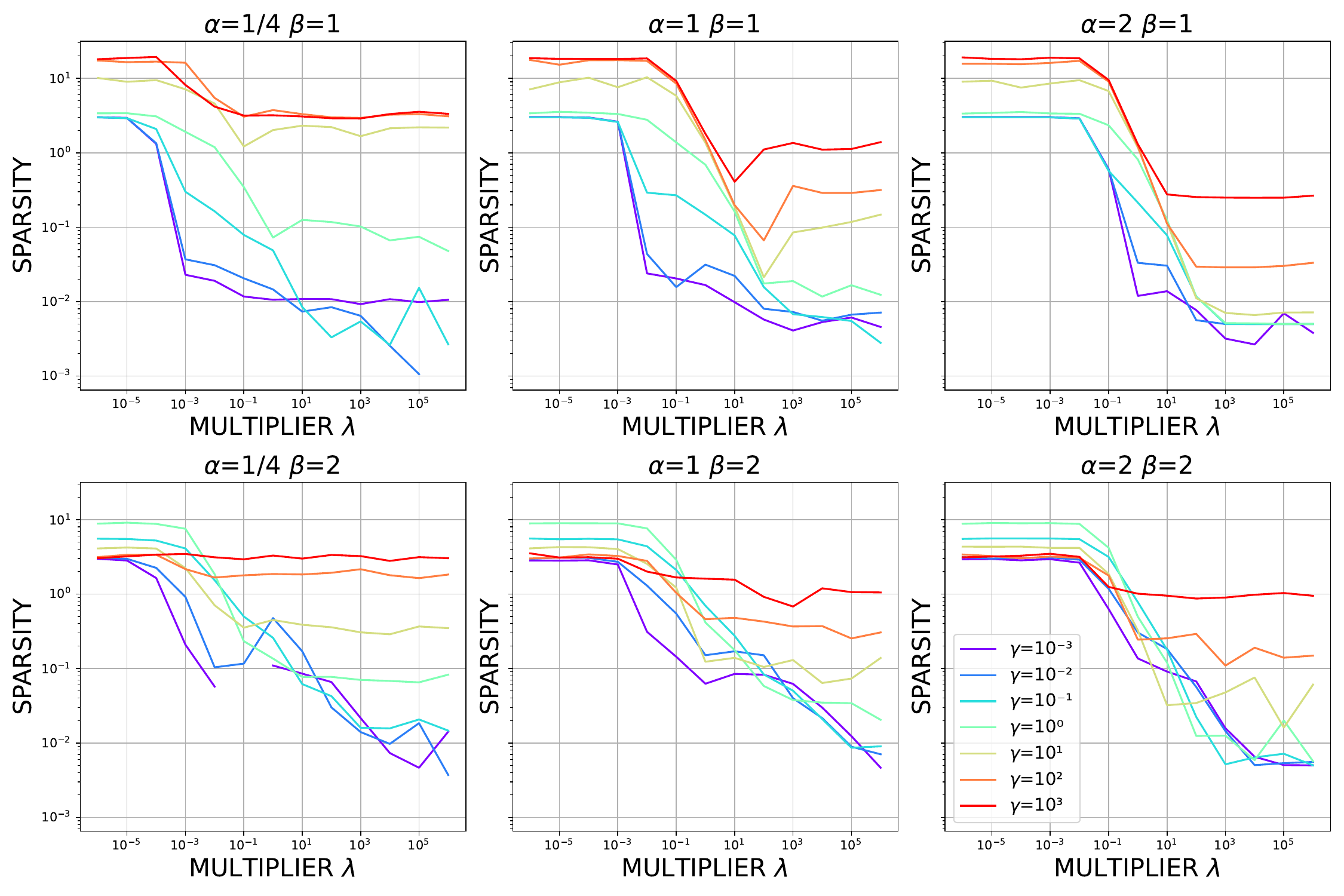}
\caption{Sparsity}
\end{subfigure}
\caption{Error distribution (a, Bhatacharyya distance between the posterior and the true distribution) and sparsity (b, $L_1$ norm of $\weight_3$)
with respect to increasing multiplier $\lambda$.
Colors indicate kernel bandwidth $\gamma$.
Each panel represents a different choice of prior order $\alpha$ and kernel order $\beta$.
}
\label{fig:mvn_survey}
\end{figure}

\fref{fig:mvn_survey} quantifies the interplay between kernel and prior over wide range of the multiplier $\lambda$ and bandwidth $\gamma$  values for selected priors $\alpha \in \{ 1/4, 1, 2 \}$ and kernels $\beta \in \{ 1, 2\}$.
Clearly, there are similar trends for all prior and kernel selections; mainly, distribution error increases and sparsity error (the opposite of sparsity) decreases with increased prior multiplier $\lambda$ i.e. there is a tradeoff between accuracy and sparsity that is dependent on the kernel width and the multiplier.
For distribution accuracy there appears to be an optimal kernel width.
Also, most trends are sigmoidal, with clustering the curves indicating insensitivity to the kernel bandwidth $\gamma$ for some choices, particularly for the parameter sparsity with the $\beta = 2$ kernel.
Note, for the following physical application, we will only use $\beta = 2$ kernel and use a threshold to reduce near zero parameters to zero (so that parameter count becomes the appropriate measure of sparsity).

\begin{figure}
\centering
\begin{subfigure}[c]{0.49\linewidth}
\includegraphics[width=0.99\linewidth]{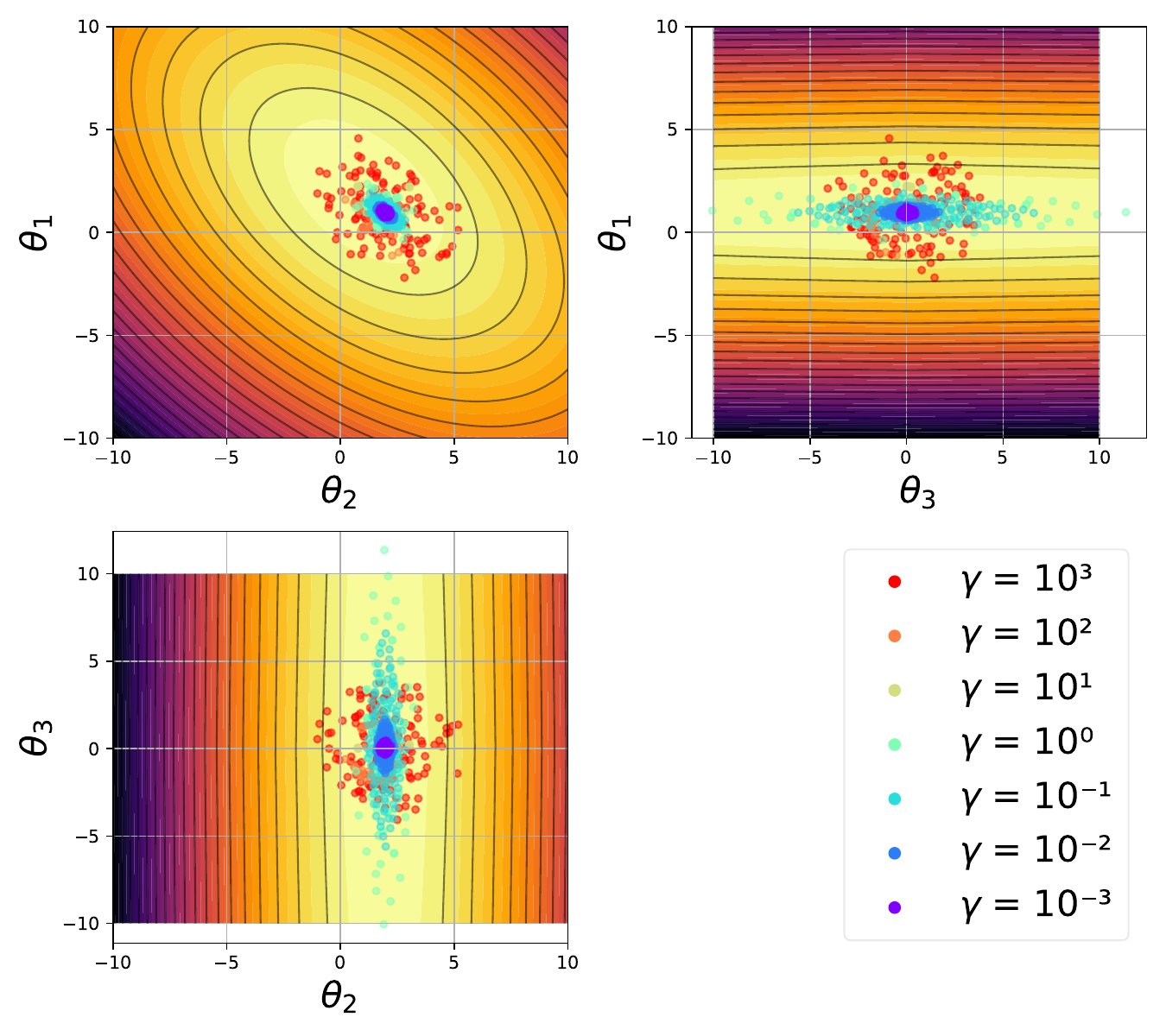}
\caption{$\lambda = 0.1$}
\end{subfigure}
\begin{subfigure}[c]{0.49\linewidth}
\includegraphics[width=0.99\linewidth]{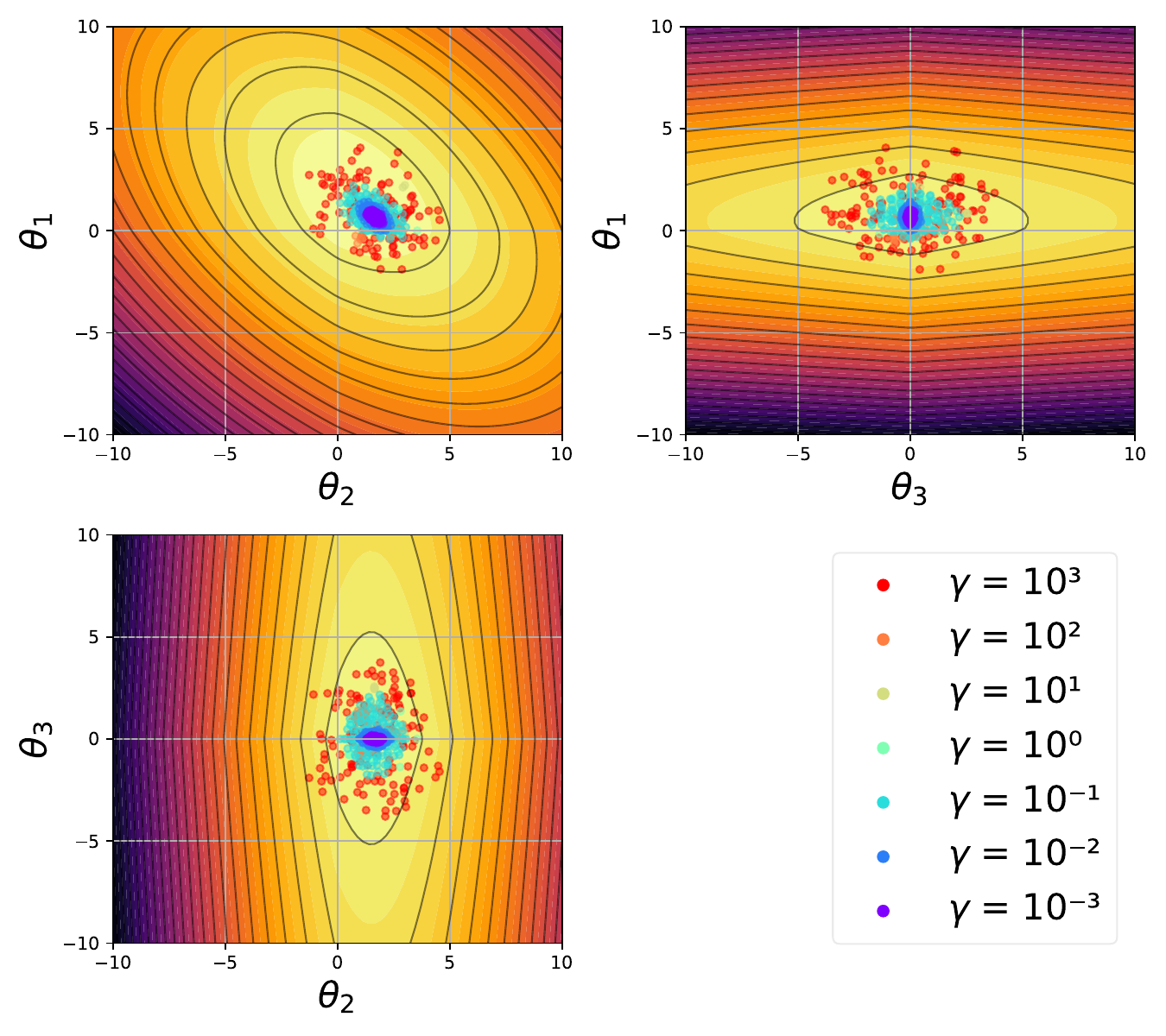}
\caption{$\lambda = 1.0$}
\end{subfigure}

\caption{
Converged particles for various bandwidths using a $\alpha = 1$ prior with multiplier $\lambda = 0.1$ (a,left) and  $\lambda = 1.0$ (b,right).
Colors indicate kernel bandwidth.
}
\label{fig:mvn_particles}
\end{figure}

Lastly, \fref{fig:mvn_particles} shows the actual arrangement of the Stein particles relative to the contours of the posterior for two values of the multiplier $\lambda$ and multiple values of the kernel bandwidth $\gamma$.
Clearly, the higher values of the bandwidth encourage the particles to explore more of the posterior landscape, while the higher value of $\lambda$ allows the particles to constrict more on the $\theta_3$ dimension, as seen by the high density of particles near $\theta_3=0$ and hence sparsify the model.

\section{Application to material constitutive modeling} \label{sec:results}

Constructing a model of the response of a hyperelastic material from data is a canonical problem in solid mechanics~\cite{fuhg2024review}.
In this case the (second Piola-Kirchhoff) stress $\stress$ determined by a potential $\potential$:
\begin{equation} \label{eq:stress_potential}
\stress = \partialb_\strain \potential
\end{equation}
so that the response is energy conservative i.e.
\begin{equation}
\oint_\strain \stress \cdot \dot{\strain} \, \dt = 0
\end{equation}
over a cycle of (the Lagrange) strain $\strain$.
To ensure a well-behaved boundary value problems based on the model \eqref{eq:stress_potential}, it is sufficient that $\potential$ is polyconvex \cite{ball1976convexity}; hence,
an input convex neural network (ICNN) is a particular FFNN that is well-suited to this problem \cite{tac2022data,chen2022polyconvex,as2022mechanics,xu2021learning,klein2022polyconvex,klein2023parametrized,kalina2024neural,fuhg2022learning,fuhg2022machine}.
Furthermore, if we assume material isotropy, the inputs of $\potential$ can be reduced to three independent scalar invariants of strain $\{I_1, I_2, I_3\}$ so that
\begin{equation} \label{eq:stress}
\stress
=  \partialb_\strain \potential
= \sum_{i = 1,2,3}
\partial_{I_i} \potential \, \partialb_{\strain} I_i
\end{equation}

\subsection{Neural Network architecture }
For this demonstration, we construct a physics-informed ICNN model that maps between three input invariants and the output stress through \eref{eq:stress}.
As in \cref{amos2017input}, a generalized ICNN network is a fully connected neural network with a skip connection and constraints on specific weights.
For an output $\outputvector = \potential$ and corresponding input $\inputvector = (I_i, i=1,2,3)$, the ICNN  with $N_\ell$ layers is the stack:
\begin{eqnarray}
\hiddenvector_1 &=& \sigma_1 \left(
\Vs_1 \inputvector +\bs_1
\right) \nonumber \\
\hiddenvector_k &=& \sigma_k \left(
\Vs_k \inputvector + \Ws_k \hiddenvector_{k-1} +\bs_k
\right)  \qquad k=2, \ldots, N_\ell-1  \label{eq:ICNN} \\
\outputvector &=& \Vs_{N_\ell} \inputvector + \Ws_{N_\ell} \hiddenvector_{N-1} +\bs_{N_\ell} \nonumber
\end{eqnarray}
with weights $\Ws_k$ and $\Vs_k$, activation functions $\sigma_k$ and $k \in [1,N_\ell]$.
The weights and biases form the set of trainable parameters $\parameters = \lbrace \Ws_k,  \Vs_k, \bs_k \rbrace$.
The output is convex with respect to the input because the weights $\Ws_k$ are constrained to be non-negative, and the activation functions $\sigma_k$ are chosen to be convex and non-decreasing \cite{amos2017input}.

As in \cref{padmanabha2024improving}, we modify the ICNN architecture to obtain better training and prediction.
For computational efficiency, we ignore bias terms and assign the same weight $\Vs_k=\Ws_k$ per layer for  $\Vs_k$ while having the weight values $\Ws_k$ to be non-negative.
We employ \emph{softplus}  as the activation functions $\sigma_k$.
Specifically, we initialize the ICNN architecture with $N_\ell = 3$ layers and $n_\ell = 30$ nodes for each layer and two hidden layers for each particle.
The model predicts the potential $\potential$, and then the stress is calculated using automatic differentiation ({\sc PyTorch} \cite{pytorch}).
We constrain the stress to be zero at the reference:
\begin{equation} \label{eq:potential_normalization}
\hat{\potential}(I_1,I_2,I_3)  = \NN(I_1,I_2,I_3) - \NN(3,3,1)-\potential_0(I_3)
\end{equation}
where $\potential_0(I_3) = n(\sqrt{I_3}-1)$ and $n$ is a constant that provides stress normalization, as in \cref{fuhg2024extreme}.

\subsection{Data Generation}
For this demonstration, we use $N_{D} = 80$ training points and $1000$ testing points, which are unseen during training.
The input space for the training and testing data is generated by uniformly sampling the deformation space $[\Fb]$:
\begin{equation}
\Fb = \Ib + \Hb \ \text{where} \ \Hb \sim \Uc[-\delta, \delta] \ .
\end{equation}
For the training data, we limit the range to $\delta = 0.2$.
Test set is formed from a parameterized deformation
\begin{equation}
\Fb =  \operatorname{diag}(1+\delta, \sqrt{1+\delta}, \sqrt{1+\delta})
\end{equation}
where $\delta \in [-0.4, 0.4]$.
Given these inputs we evaluate the same truth model as in our previous work \cref{padmanabha2024improving}
\begin{equation}
\Psi(I_1,I_2,J) = -\frac{\vartheta_{1}}{2} J_{m} \log \left( 1 - \frac{I_{1}-3}{J_{m}} \right) - \vartheta_{2} \log \left( \frac{I_{2}}{J} \right) + \vartheta_{3} \left( \frac{1}{2} (J^{2}-1) - \log J \right) ,
\end{equation}
with $J=\sqrt{I_{3}}$, $J_{m}= 77.931$, $\vartheta_{1} =2.4195$,  $\vartheta_{2} =-0.75$ and $\vartheta_{3} =1.20975 $, and introduce multiplicative noise to the outputs to generate the input-output pairs
\begin{equation}
\data = \left( \strain_i, \varsigma \epsilonb \stress_i \right)
\ \text{where} \ \epsilonb \sim \Nc(\mathbf{1}, \Ib)
\label{eq:datasamples}
\end{equation}
Here $\varsigma$ acts as a noise level.

\subsection{Performance evaluation}

Pushforward posteriors can be generated from the Stein ensemble of realizations $\{ \parameters_b \}_1^{N_r}$ via
\begin{equation}
\stress_a = \hat{\stress}(\strain; \parameters_a)  \ \text{where} \ \parameters_a \in \{ \parameters_b \}_{b=1}^{N_r}
\end{equation}
We use the Wasserstein-1 ($\mathcal{W}_1$) distance to compare the push-forward posteriors $\prob(\stress) = \sum_a \delta(\stress - \stress_a)$:
\begin{equation}
\mathcal{W}_1(\prob(\hat{\stress}), \prob(\stress))
= \int \vert \CDF(\hat{\stress}) - \CDF(\stress)\vert \, \mathrm{d}\stress
\end{equation}
where the cumulative distribution functions (CDFs) associated with the probability density functions $\prob(\stress)$ are defined empirically from the particle samples pushed forward through the model and data, respectively.
Being a distance, a smaller $\mathcal{W}_1$ implies that the two distributions are more similar.
We apply the $\mathcal{W}_1$ distance to the predictions at every time and also produce a single similarity measure by summing the values across all output times.

\subsection{Graph condense}
\begin{figure}
\centering
\includegraphics[width=0.95\linewidth]{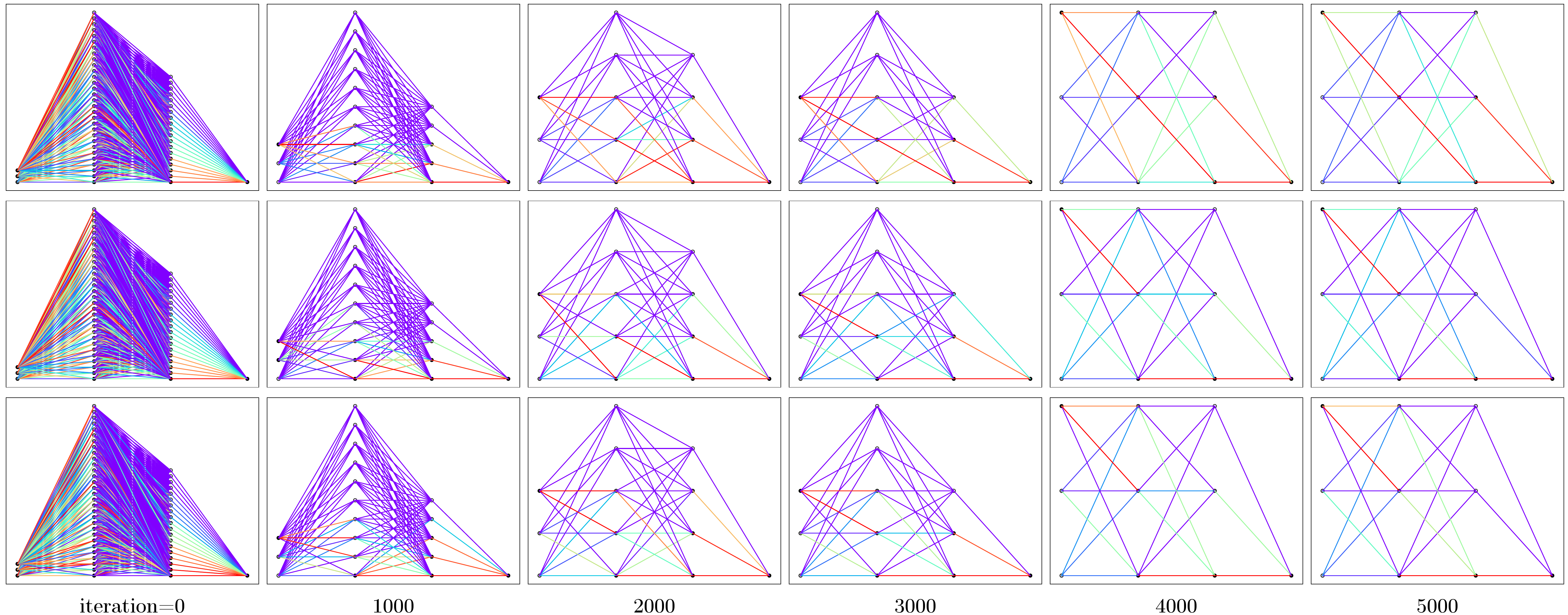}
\caption{
Sequence of graphs for 3 randomly selected particles during the iteration process.
Initial models are on the left, and the final on the right.
Nodes are colored by importance \eref{eq:importance}, and edges are colored by weight $W_{ij}$.
All plotted values are normalized per layer.
}
\label{fig:graph_condense_sequence}
\end{figure}

\fref{fig:graph_condense_sequence} illustrates how the graphs representing the ensemble of FFNNs evolve under the Stein flow for $\alpha=1/4$ prior with multiplier $\lambda=0.05$.
The bandwidth for the $\beta=2$ kernel was adapted using $\gamma = \sqrt{1/2 \bar{d} / \log(N_r+1)}$ where $\bar{d}$ is the median distance between particles.
Three randomly selected particles in an ensemble of $N_r = 10$ are shown over a sequence of iterations progressing from left, where the models are clearly complex, to the right, where they are evidently sparse.
All particles are distinct but become more similar as the gradient descent and graph condensation iterations progress.
It is apparent that the method is sampling the diversity of likely parameterizations while removing permutational variations that artificially increase parameter variance.
Note that any particular realization/particle can become more complex as it can fill out common graph for ensemble during a stage of the SVGD.

\begin{figure}
\centering
\includegraphics[width=0.95\linewidth]{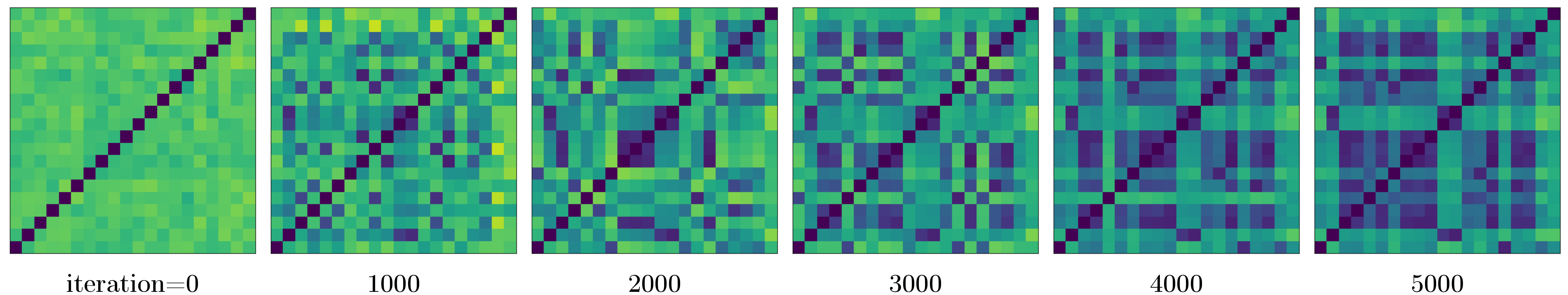}
\caption{Evolution of graph distances of the ensemble during SVGD (blue pixels indicate zero and yellow pixels indicate maximum value).
Note this illustration uses $N_r=20$ particles.}
\label{fig:graph_distance_evolution}
\end{figure}

\fref{fig:graph_distance_evolution} illustrates the evolution of interparticle distances for a $N_r=20$ ensemble, as calculated by \eref{eq:distances}.
Since the initial weights are randomized, the particles start out effectively equidistant from each other.
Then we see that the particles have a transient where they attain a maximum dispersion.
Eventually through the SVGD with graph condensation we see that subsets of particles grow close to each other but there is still diversity in the parameterizations.

\begin{figure}
\centering
\begin{subfigure}[c]{0.32\linewidth}
\includegraphics[width=0.99\linewidth]{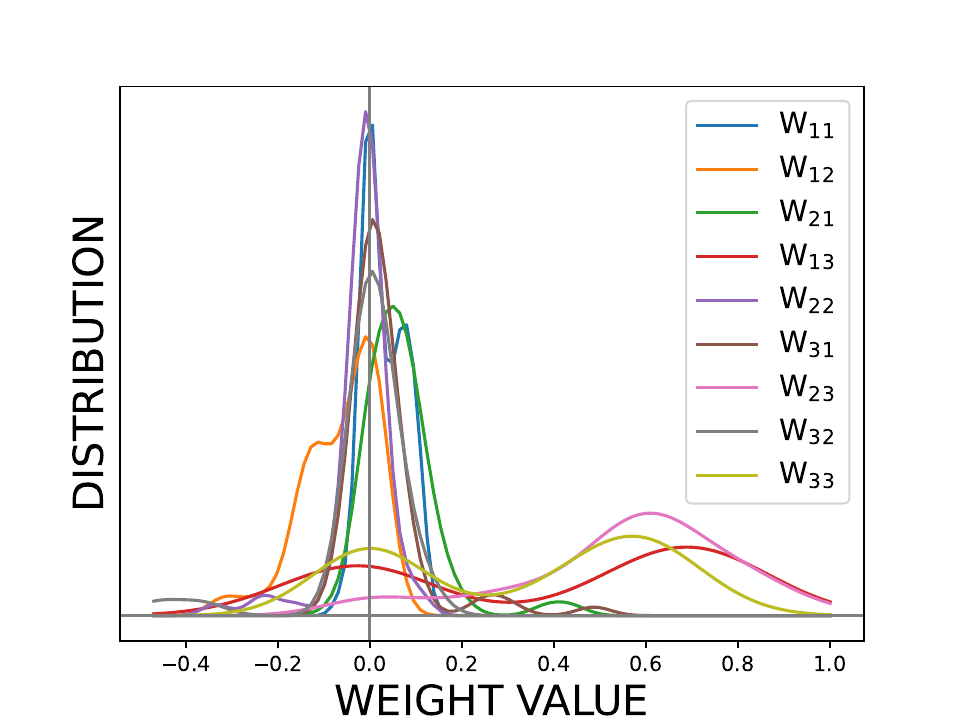}
\caption{$\Ws_0$}
\end{subfigure}
\begin{subfigure}[c]{0.32\linewidth}
\includegraphics[width=0.99\linewidth]{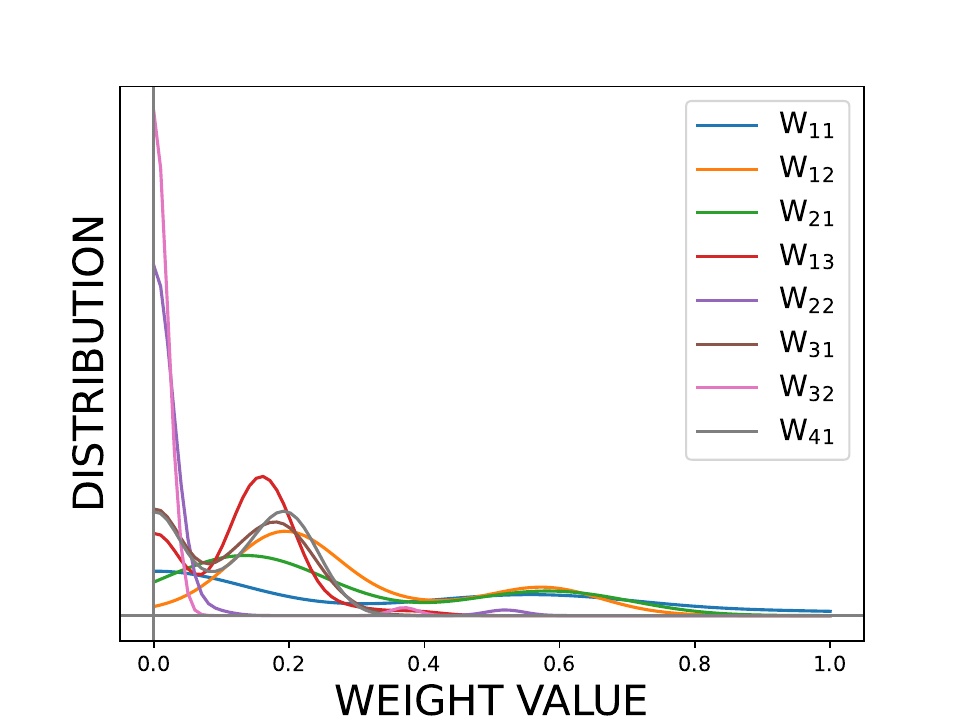}
\caption{$\Ws_1$}
\end{subfigure}
\begin{subfigure}[c]{0.32\linewidth}
\includegraphics[width=0.99\linewidth]{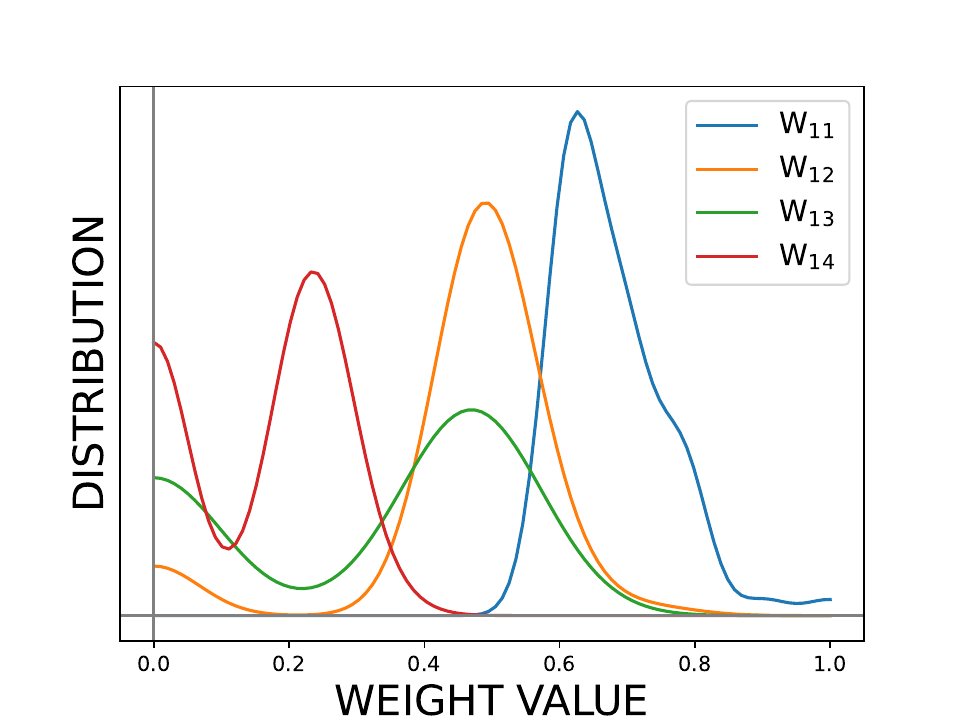}
\caption{$\Ws_2$}
\end{subfigure}

\caption{Final kernel density estimates (KDE) of the distribution of weights:
(a) $\Ws_0$, (b) $\Ws_1$, (c) $\Ws_2$.
Weights $\Ws_k$ connect layers $\ell=k$ and $\ell=k+1$ and $[ \Ws_k ]_{ij}$ are the components of $\Ws_k$ which are abbreviated as $W_{ij}$ in the legends of the plots.
Note narrow distributions centered at zero are omitted for clarity.
}
\label{fig:weight_distributions}
\end{figure}

\fref{fig:weight_distributions} shows the final distributions of $\Ws_\ell$, $\ell \in 0,1,2$ using kernel density estimation (KDE) of the particle samples.
The input to first hidden layer weight matrix $\Ws_0$ is not constrained to be positive by the ICNN formulation and has number of weight values that are widely centered at zero and a few that have distinct positive peaks.
The middle weight matrix $\Ws_2$ has a mixture of zero and non-zero centered distributions indicating the connections in between the two hidden layers are sparse.
Interestingly, the last hidden layer to output weight matrix $\Ws_2$ has strongly ordered weights, although some still have significant zero centered modes.
The fact that most distributions are unimodal or bimodal with one peak at zero indicates the importance ranking is effective in aligning the weights; however, there are likely better metrics to sort and align the parameterizations.

\begin{figure}
\centering
\begin{subfigure}[c]{0.49\linewidth}
\includegraphics[width=0.99\linewidth]{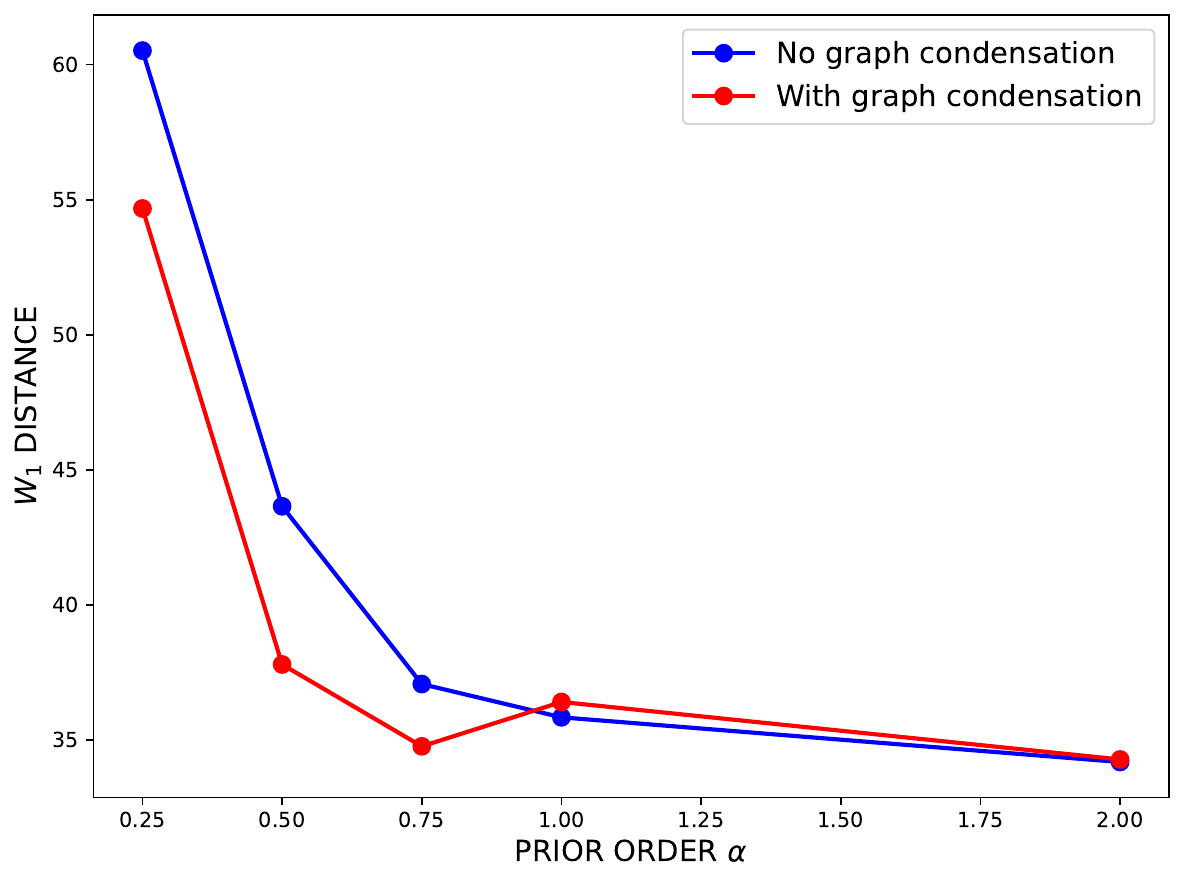}
\caption{Accuracy}
\end{subfigure}
\begin{subfigure}[c]{0.49\linewidth}
\includegraphics[width=0.99\linewidth]{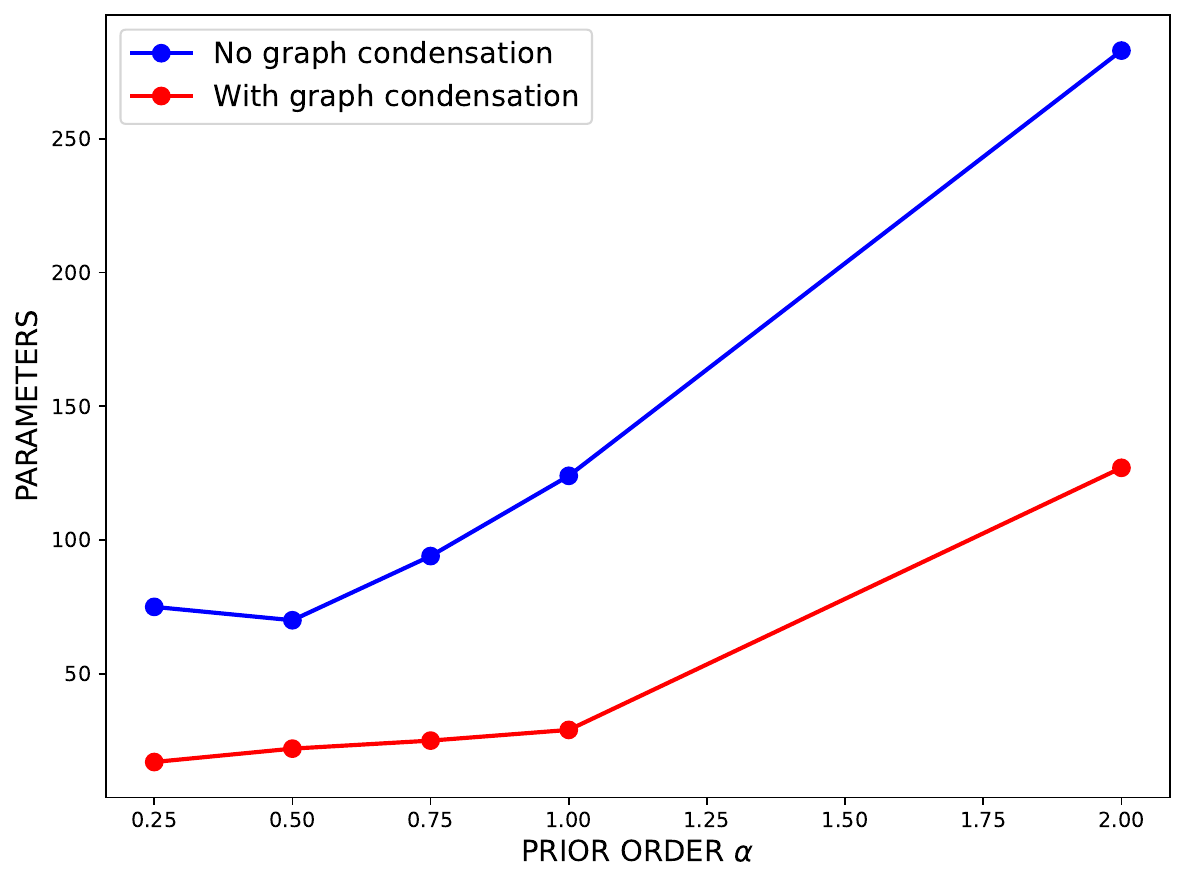}
\caption{Sparsity}
\end{subfigure}
\caption{
Accuracy (a, $\mathcal{W}_1$ distance) and  sparsity (b, parameter count) for various priors with and without graph condensation.
Note initial parameter count is 1020.
}
\label{fig:graph_condense_speedup}
\end{figure}

Lastly \fref{fig:graph_condense_speedup} provides evidence that a sparsifying SVGD, shown in \Aref{alg:svgd}, with graph condense, presented in \Aref{alg:graph_condense}, is more effective than without graph condensation at reducing model complexity across a wide range of prior choices.
This comes with no reduction in accuracy and an attendant speed-up that is proportional to the complexity reduction.
We found that the computational cost of an iteration is roughly proportional to the parameter count of the models in the Stein ensemble, hence, as the models simplify the gradient descent iterations speed up.
Without graph condense, near-zero parameters are allowed to re-emerge, whereas with graph condensation the particles in the subsequent stages of the SVGD work within a restricted weight graph / template.

\subsection{Convergence}
We explored the sensitivity of our method's performance with respect to various hyperparameters.

\begin{figure}
\centering
\begin{subfigure}[c]{0.49\linewidth}
\includegraphics[width=0.99\linewidth]{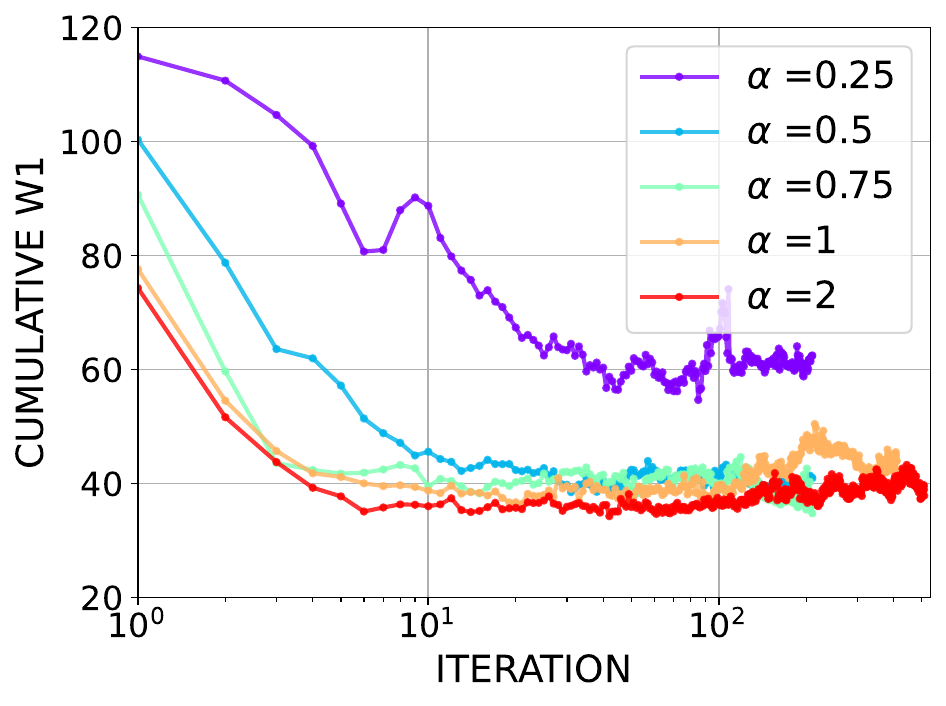}
\caption{Accuracy}
\end{subfigure}
\begin{subfigure}[c]{0.49\linewidth}
\includegraphics[width=0.99\linewidth]{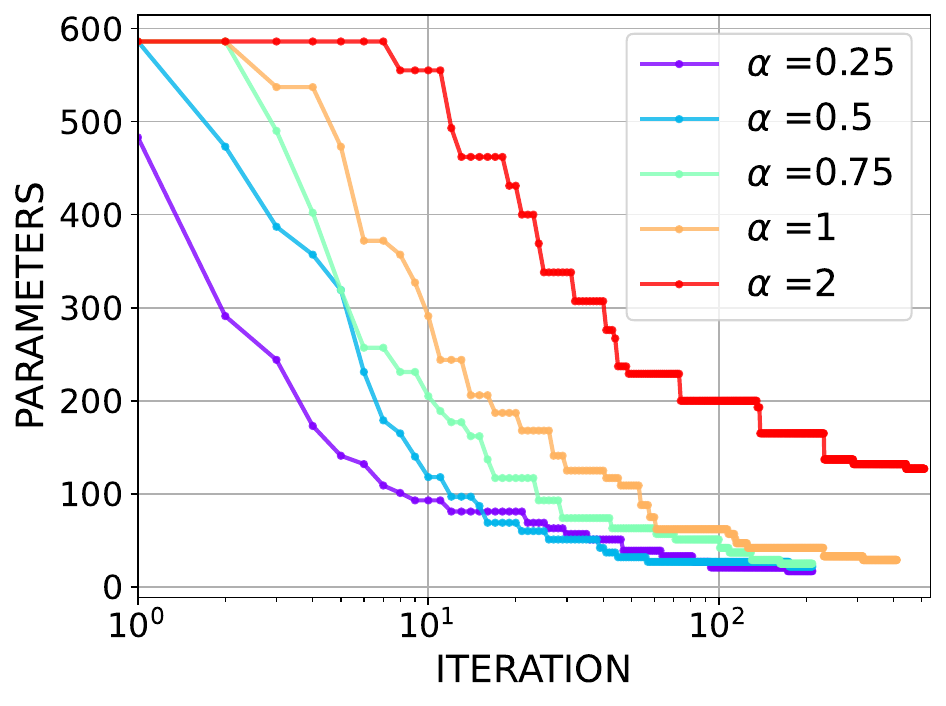}
\caption{Sparsity}
\end{subfigure}
\caption{
Convergence of accuracy (a) and sparsity (b) for various prior orders $\alpha$ with 10\% noise.
Note initial parameter count for all models is 1020.
}
\label{fig:convergence_priors}
\end{figure}

\fref{fig:convergence_priors} compares the accuracy, as measured by the $\mathcal{W}_1$ distance, and sparsity, in terms of parameter count, of the Stein ensemble over the gradient descent iterations for various prior orders $\alpha$ for $\varsigma = 10\%$ (the noise level in \eref{eq:datasamples}).
As expected, the trends in accuracy and complexity are complementary.
The most complex models are also the most accurate; however, the behavior for all priors $\alpha \leq 1$ is similar.
The $L_2$ prior ($\alpha = 2$) is an outlier since it is generally not considered a prior that strongly induces parameter sparsity (it does force parameters to zero but has no corners in its level sets), as evidenced by the flat stages in the parameter counts.
However, there are abrupt decreases in parameter counts when the graph condensation is applied based on a weight tolerance.
This effect in sparsification is apparent with the higher $\alpha$ priors and become more smooht for the lower $\alpha$ priors.
For all priors, we observe that as $\alpha \to 0$, sparsity is induced earlier in the iteration process, but this also delays the error convergence.
The influence of $\alpha$ on model complexity is evident from the final number of active weights at the last iteration: for $\alpha = [0.25, 0.5, 0.75, 1, 2]$, the corresponding active weights are $[17, 22, 25, 29, 127]$.
This highlights how smaller values of $\alpha$ lead to stronger sparsity, whereas larger values result in more complex models with more active parameters.
Consequently, as $\alpha \to 0$, computational time decreases due to the algorithm operating on a smaller/condensed graph with fewer active weights, whereas priors $\alpha = 1$ and $\alpha = 2$ require more iterations to converge compared to $\alpha < 1$, as they involve training a larger number of active weights.
For this case study, we consider an ensemble of $10$ particles.
The choice of $\alpha$ enables the algorithm to explore the Pareto front between accuracy and sparsity objectives, balancing the trade-offs inherent to these competing goals.

\begin{figure}
\centering
\begin{subfigure}[c]{0.49\linewidth}
\includegraphics[width=0.99\linewidth]{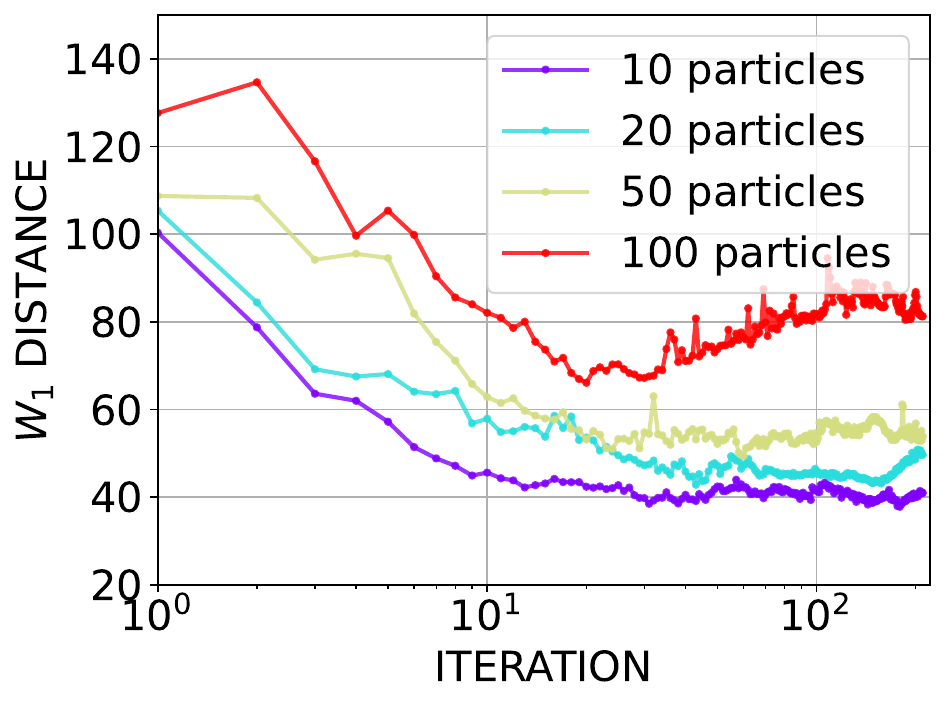}
\caption{Accuracy}
\end{subfigure}
\begin{subfigure}[c]{0.49\linewidth}
\includegraphics[width=0.99\linewidth]{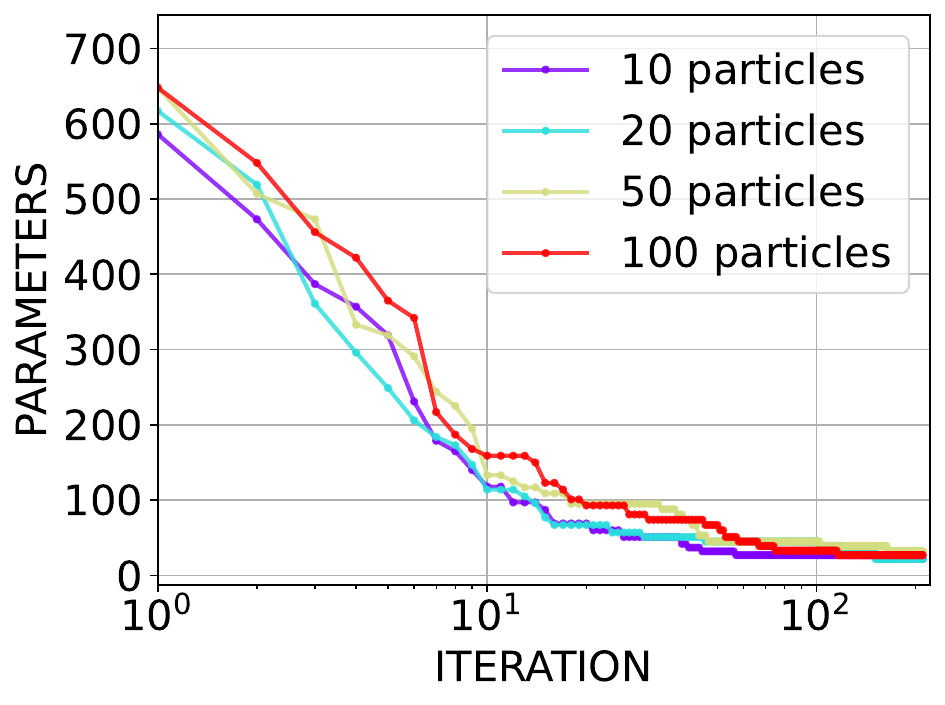}
\caption{Sparsity}
\end{subfigure}
\caption{
Convergence of accuracy (a) and sparsity (b) for various ensemble sizes $N_r$ with 10\% noise.
Note initial parameter count for all models is 1020.
}
\label{fig:convergence_ensemble}
\end{figure}

\fref{fig:convergence_ensemble} compares the convergence of the Stein ensemble for various sizes $N_r$.
Sparsity is relatively insensitive to the ensemble size, while accuracy is noticeably affected.
In this analysis, where we consider the prior $\alpha = 0.5$ and noise level of 10\%, we observe that for 100 particles, there is some evidence of overfitting or crowding behavior, likely due to the repulsive kernels not allowing all particles to occupy the high-probability region of the posterior.
Allowing the kernel width to vary adaptively may reduce this sensitivity.
Nevertheless, for $N_r \le 50$, the results are fairly robust, yielding final model sizes on the order of 10 parameters.

\begin{figure}
\centering
\begin{subfigure}[c]{0.49\linewidth}
\includegraphics[width=0.99\linewidth]{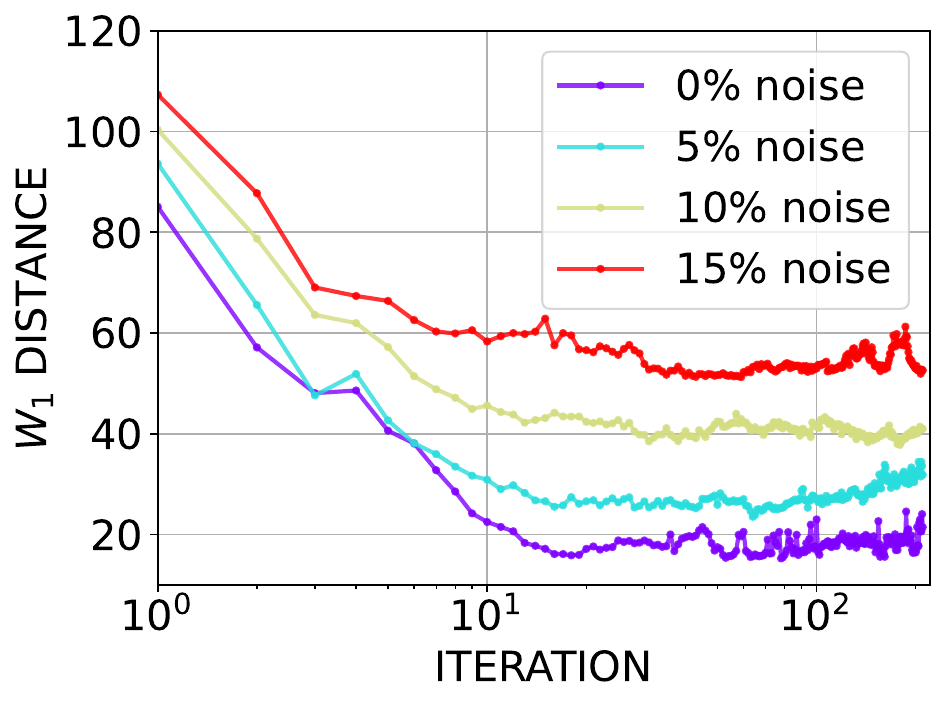}
\caption{Accuracy}
\end{subfigure}
\begin{subfigure}[c]{0.49\linewidth}
\includegraphics[width=0.99\linewidth]{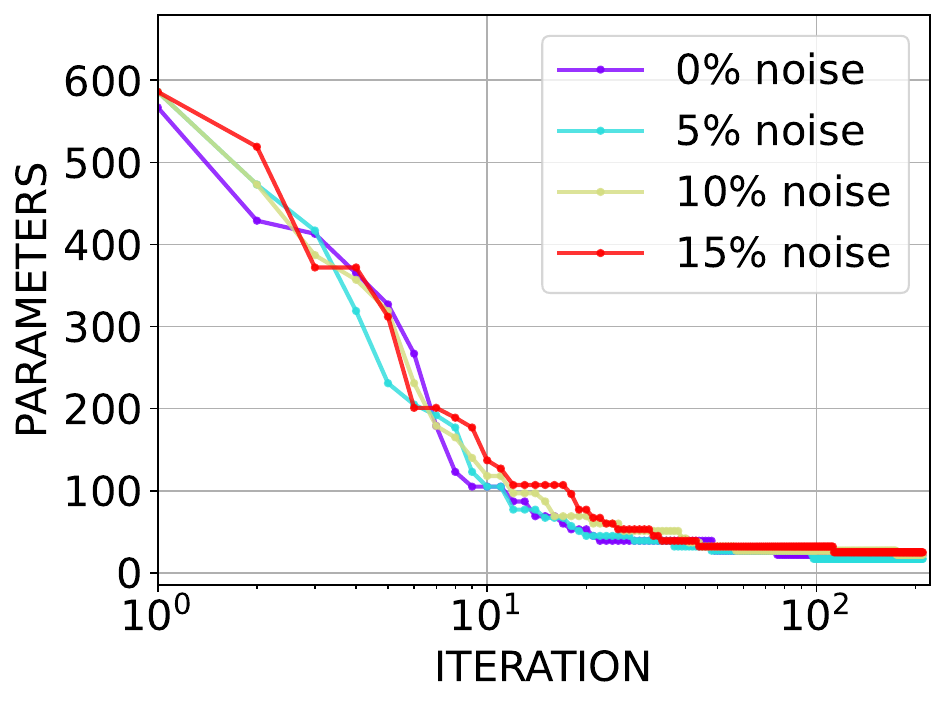}
\caption{Sparsity}
\end{subfigure}
\caption{
Convergence of accuracy (a) and sparsity (b) for various noise levels $\varsigma$
Note initial parameter count for all models is 1020. We consider the prior $\alpha = 0.5$ and number of particles $10$.}

\label{fig:convergence_noise}
\end{figure}

\begin{figure}
\centering
\begin{subfigure}[c]{0.59\linewidth}
\includegraphics[width=0.99\linewidth]{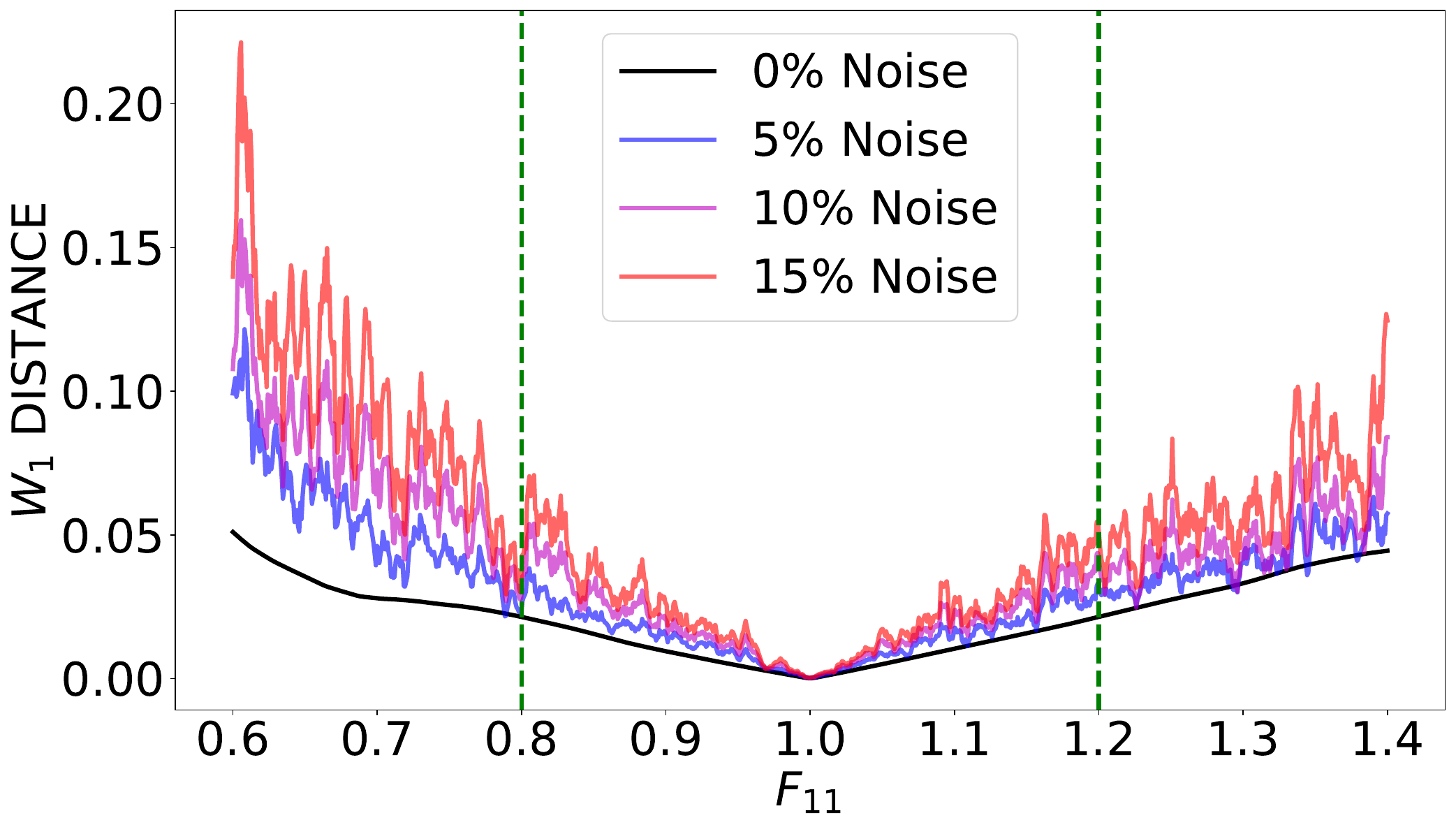}
\caption{Noise levels}
\end{subfigure}
\begin{subfigure}[c]{0.59\linewidth}
\includegraphics[width=0.99\linewidth]{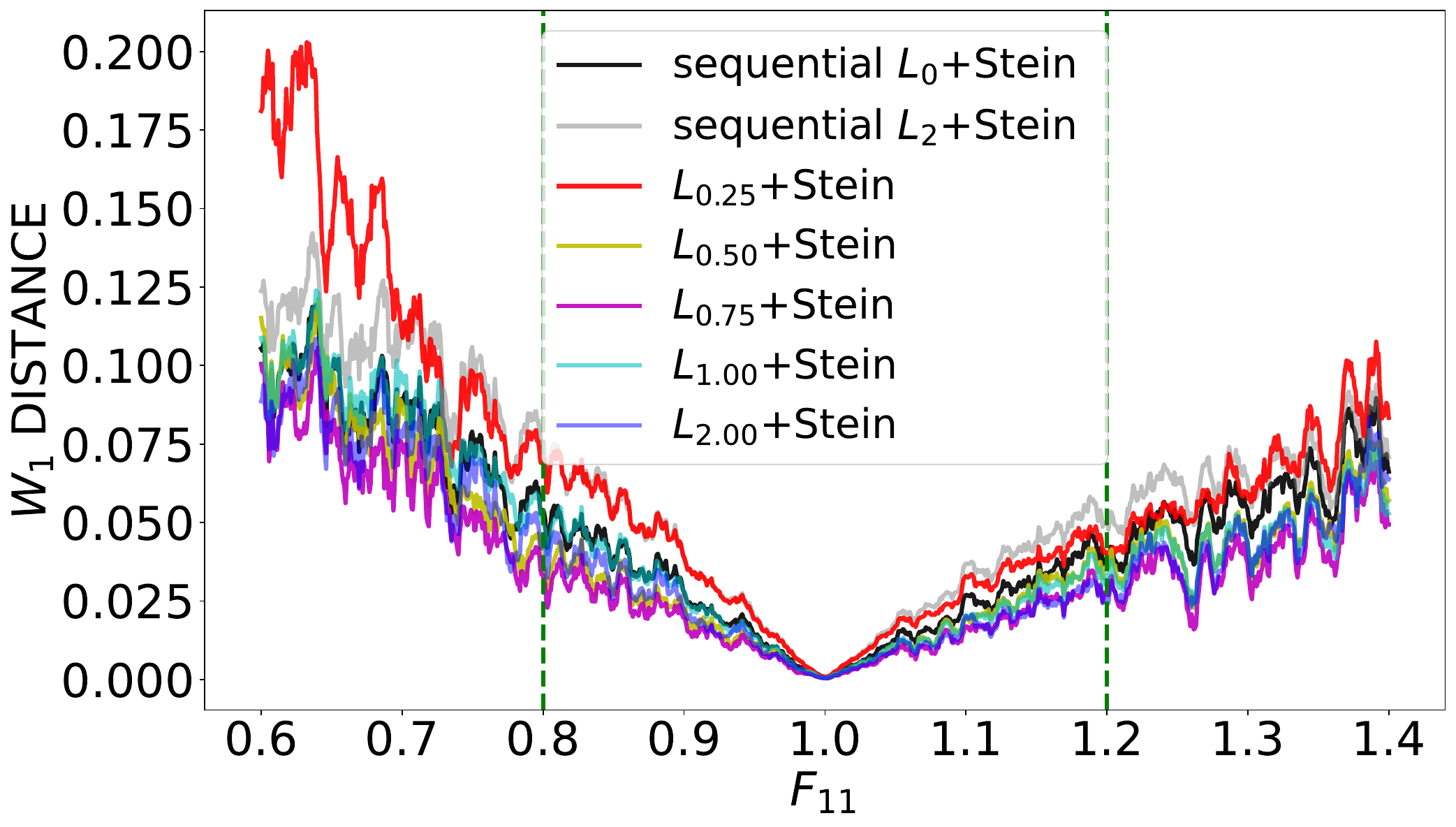}
\caption{Method comparison}
\end{subfigure}
\caption{Pushforward errors as measured by the $\mathcal{W}_1$ distance as a function of deformation $F_{11}$: (a) for various multiplicative noise levels, and (b) comparison of methods for 10\% noise level. In all cases, the model was initialized with 1,020 parameters.
Note the high-frequency noise in the $\mathcal{W}_1$ curves has been filtered with a moving average to help distinguish the trends.}

\label{fig:noisy_pushforwards}
\end{figure}

\fref{fig:convergence_noise} compares convergence for increasing noise levels $\varsigma$.
Again sparsity is relatively insensitive to increased noise, but accuracy suffers, as expected.
The errors increase approximately linearly with multiplicative noise.
\fref{fig:noisy_pushforwards}a provides a detailed view of the $\mathcal{W}_1$ distances as a function of the input, deformation $F_{11}$, and noise.
The errors are zero at $F_{11}=0$ due to multiplicative noise and model construction, \eref{eq:potential_normalization}, and generally increase linearly away from $F_{11}=0$.
\fref{fig:noisy_pushforwards}b shows that the concurrent SVGD method proposed in this work can outperform the sequential sparsify and then propagate a Stein ensemble developed in our previous work \cite{padmanabha2024improving}.
In all cases, the model was initialized with the same number of weights: $1020$ parameters. The effect of graph condensation significantly enhances the model's efficiency. In our previous work~\cite{padmanabha2024improving}, without graph condensation, the $L_{2}+$ Stein sequential method utilized $1005$ parameters. In contrast, graph condensation in this work reduces the number of active parameters to $127$, achieving similar accuracy while significantly decreasing computational time. Hence, operating on a condensed graph with fewer active weights allows the algorithm to achieve lower computational complexity and improved efficiency.

All these studies were performed with a fixed penalty $\lambda=0.05$.
This was done for simplicity and clarity in the comparison of trends.
Since there is an optimal $\lambda$ for each choice of the prior order, kernel order, and size, as well as trade-offs between accuracy and complexity, in the next section, we demonstrate the effectiveness of using an adaptive penalty scheme. For all the cases, including the adaptive case, a threshold of $10^{-3}$ is employed to exclude negligible weights, thereby improving computational efficiency. The neural network models in this work were trained on a single NVIDIA RTX A$6000$ GPU.

\subsection{Adaptive penalty}
We assume the existence of an optimal kernel and prior in terms of a particular balance of sparsity and accuracy.
In the previous demonstrations, we have presented a limited exploration focusing primarily on trends with changes in the hyperparameters.
In this section, we compare the empirical study of the penalty $\lambda$ to a control scheme that starts with a low $\lambda = 0.01$ so as not to strongly influence accuracy and then increases $\lambda$ till accuracy suffers and finally reverts $\lambda$ to the initial value with the sparsification fixed by the graph condensation after $4000$ iterations. In this analysis, we fix $\alpha$ at $0.5$ and number of particles to be $10$.

\begin{figure}
\centering
\begin{subfigure}[b]{0.49\linewidth}
\includegraphics[width=0.99\linewidth]{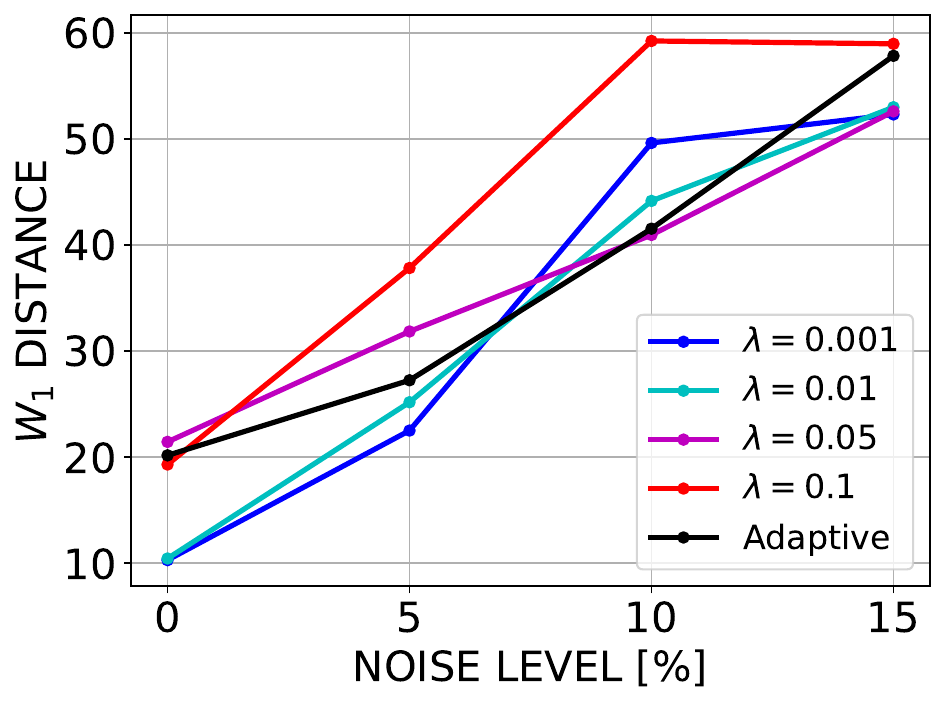}
\caption{Accuracy}
\end{subfigure}
\begin{subfigure}[b]{0.49\linewidth}
\includegraphics[width=0.99\linewidth]{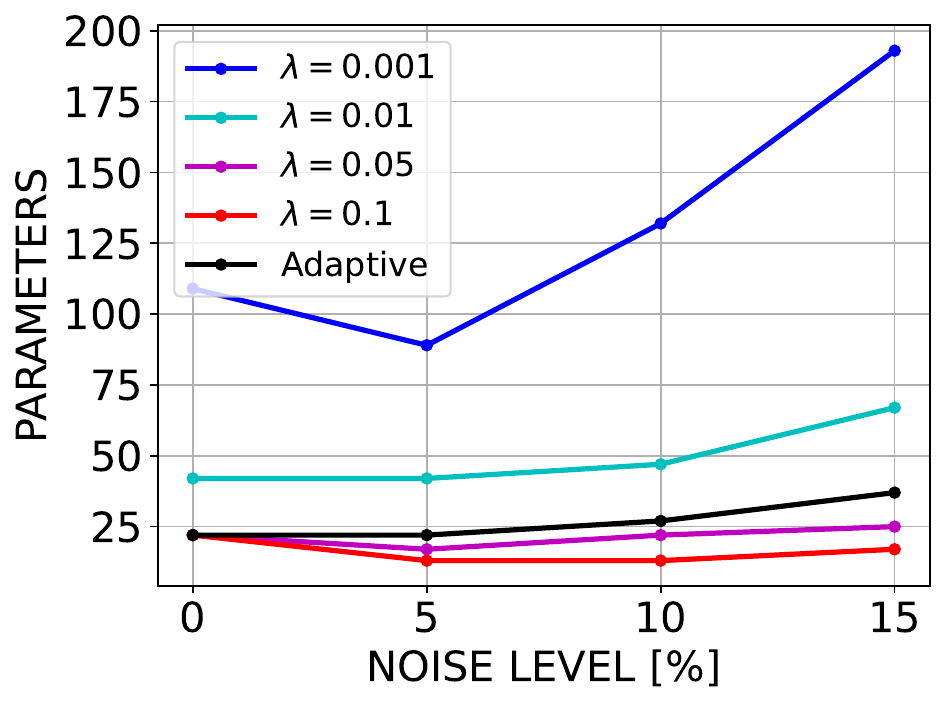}
\caption{Sparsity}
\end{subfigure}
\caption{ Comparison of (a) accuracy and (b) sparsity for fixed and adaptive penalty. }
\label{fig:adaptive_penalty}
\end{figure}

\fref{fig:adaptive_penalty} compares the performance of a higher than optimal $\lambda = 0.1 $, a near optimal $\lambda = 0.05$ and lower than optimal penalty $\lambda=0.01$ with the adaptive penalty scheme outlined in \Aref{alg:svgd}.
Clearly, the simple adaptive scheme promotes both accuracy and sparsity that is near optimal across a wide range of noise levels $\varsigma$.

\section{Conclusion} \label{sec:conclusion}

We developed an enhancement of the Stein variational gradient descent scheme that reduces the complexity of the models in the ensemble via a sparsifying prior.
It is based a parameter alignment and condensation procedure ameliorates the potential for spurious particle repulsion in the Stein ensemble.
The balance between accuracy and complexity goals is tunable via the choice of the penalty parameter and the prior order.
Generally, the performance is not overly sensitive to these choices.
We also observed some performance dependence on the repulsive kernel shape and size.
Significantly, the reduction in complexity leads to iteration speed up due to the descent operating in a smaller parameter space, whereby the extra cost of the graph sorting and reconfiguring is more than offset by this efficiency.

In future work, we intend to explore other sorting criteria in the graph condense algorithm and  replacing the $L_\alpha$ like priors with smoother Huber-like priors \cite{huber1992robust} to possibly improve convergence.
Also, more sophisticated penalty $\lambda$ and kernel width $\gamma$ adaptation will likely improve performance.
We will apply the method to other neural network architectures, particularly large, multicomponent models, where the graph condensation can be done on components of the neural network, either in sequence or guided by expert knowledge or some model selection scheme.
For the the large-scale deployment, we also foresee the need for optimization of the data structures and algorithms for efficient sorting and model reconfiguration.

\section*{Acknowledgements}
GAP and NB were supported by the SciAI Center, and funded by the Office of Naval Research (ONR), under Grant Number N00014-23-1-2729.
REJ and CS were supported by the U.S. Department of Energy, Advanced Scientific Computing program.
CS was also supported by the Scientific Discovery through Advanced Computing (SciDAC) program through the FASTMath Institute.

Sandia National Laboratories is a multi-mission laboratory managed and operated by National Technology \& Engineering Solutions of Sandia, LLC (NTESS), a wholly owned subsidiary of Honeywell International Inc., for the U.S. Department of Energy’s National Nuclear Security Administration (DOE/NNSA) under contract DE-NA0003525. This written work is authored by an employee of NTESS. The employee, not NTESS, owns the right, title and interest in and to the written work and is responsible for its contents. Any subjective views or opinions that might be expressed in the written work do not necessarily represent the views of the U.S. Government. The publisher acknowledges that the U.S. Government retains a non-exclusive, paid-up, irrevocable, world-wide license to publish or reproduce the published form of this written work or allow others to do so, for U.S. Government purposes. The DOE will provide public access to results of federally sponsored research in accordance with the DOE Public Access Plan.

This research utilized the Delta advanced computing and data resource, which is supported by the National Science Foundation (award OAC 2005572) and the State of Illinois. Delta is a collaborative initiative between the University of Illinois Urbana-Champaign and its National Center for Supercomputing Applications (NCSA). This work leveraged the Delta system at NCSA through allocation MTH240015, provided by the Advanced Cyberinfrastructure Coordination Ecosystem: Services $\&$ Support (ACCESS) program, supported by National Science Foundation grants \#2138259, \#2138286, \#2138307, \#2137603, and \#2138296.


\end{document}